\PassOptionsToPackage{prologue,dvipsnames}{xcolor}
\documentclass[10pt,twocolumn,letterpaper]{article}

\usepackage[pagenumbers]{cvpr} 

\definecolor{cvprblue}{rgb}{0.21,0.49,0.74}
\usepackage[pagebackref,breaklinks,colorlinks,allcolors=cvprblue]{hyperref}

\usepackage{url}
\usepackage{booktabs}
\usepackage{adjustbox}
\usepackage{multicol}
\usepackage{multirow}
\usepackage{colortbl}
\usepackage{caption}
\usepackage{subcaption}
\usepackage{floatrow}
\usepackage{pgfplots}
\usepackage[accsupp]{axessibility}  

\def\paperID{23776} 
\def\confName{CVPR}
\def\confYear{2026}

\title{Reviving ConvNeXt for Efficient Convolutional Diffusion Models}

\author{
\begin{tabular}{cccc}
Taesung Kwon\textsuperscript{1,2} &
Lorenzo Bianchi\textsuperscript{2,3,4} &
Lennart Wittke\textsuperscript{2} &
Felix Watine\textsuperscript{2} \\
Fabio Carrara\textsuperscript{3} &
Jong Chul Ye\textsuperscript{1} &
Romann Weber\textsuperscript{$\dagger$} &
Vinicius Azevedo\textsuperscript{$\dagger$}
\end{tabular}
\\[1.2em]
{\small \textsuperscript{1}KAIST} \hspace{25pt}
{\small \textsuperscript{2}ETH Z\"{u}rich} \hspace{25pt}
{\small \textsuperscript{3}ISTI-CNR} \hspace{25pt}
{\small \textsuperscript{4}University of Pisa}
}

\begin{document}

\twocolumn[{%
\renewcommand\twocolumn[1][]{#1}%
\maketitle
\begin{center}
    \vspace{-0.5cm}
    \centering
    \captionsetup{type=figure}
    \includegraphics[width=0.8\linewidth]{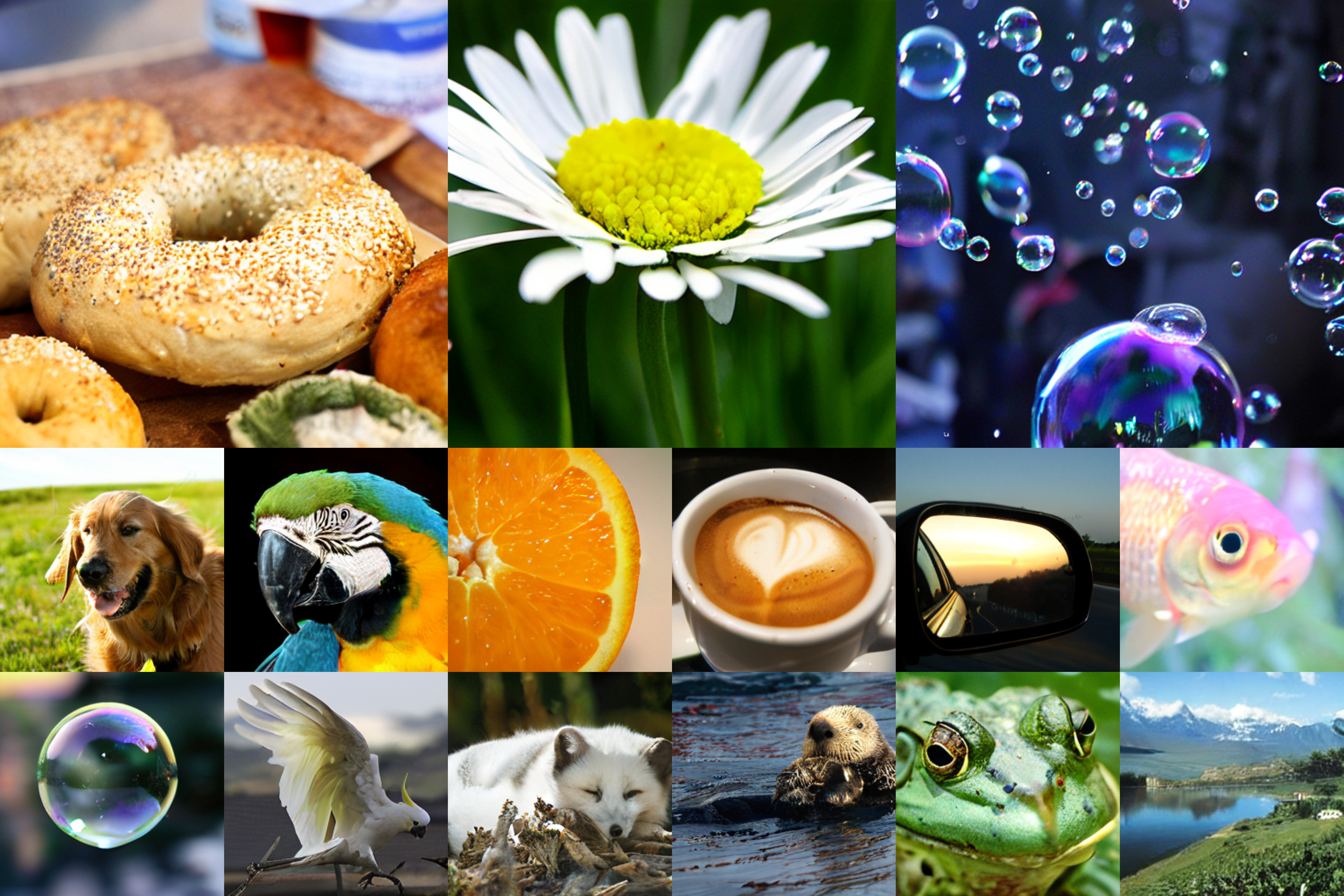}
\vspace{-0.1cm}
\caption{\textbf{Diffusion models with fully convolutional backbones achieve high-quality image generation with state-of-the-art efficiency.} We show selected samples from two of our class-conditional FCDM-XL models trained on ImageNet at 512$\times$512 and 256$\times$256 resolution.}
\label{fig:main}
\end{center}}]

\let\thefootnote\relax\footnotetext{\textsuperscript{$\dagger$}Independent researcher.}

\begin{abstract}
Recent diffusion models increasingly favor Transformer backbones, motivated by the remarkable scalability of fully attentional architectures. Yet the locality bias, parameter efficiency, and hardware friendliness—the attributes that established ConvNets as the efficient vision backbone—have seen limited exploration in modern generative modeling. Here we introduce the fully convolutional diffusion model (FCDM), a model having a backbone similar to ConvNeXt, but designed for conditional diffusion modeling. We find that using only 50$\%$ of the FLOPs of DiT-XL/2, FCDM-XL achieves competitive performance with 7$\times$ and 7.5$\times$ fewer training steps at 256$\times$256 and 512$\times$512 resolutions, respectively. Remarkably, FCDM-XL can be trained on a 4-GPU system, highlighting the exceptional training efficiency of our architecture. Our results demonstrate that modern convolutional designs provide a competitive and highly efficient alternative for scaling diffusion models, reviving ConvNeXt as a simple yet powerful building block for efficient generative modeling.
\end{abstract}

\let\thefootnote\relax\footnotetext{Official implementation is available \href{https://github.com/star-kwon/FCDM}{\textcolor{magenta}{here}}.}

\section{Introduction}
\label{sec:intro}
Over the past decade, convolutional neural networks (ConvNets)~\citep{lecun1989backpropagation, krizhevsky2012imagenet, simonyan2015very, szegedy2015going, ronneberger2015u, he2016deep, xie2017aggregated, huang2017densely, howard2017mobilenets, tan2019efficientnet} have driven most major advances in computer vision. Their success stems in part from the implicit “sliding window” mechanism, which embeds a strong locality inductive bias and enables learning effective visual representations with far fewer parameters than fully connected layers.
With the incorporation of patch embeddings in the Vision Transformer (ViT)~\citep{dosovitskiy2021an, liu2021swin}, Transformers~\citep{vaswani2017attention} began to be actively explored in computer vision as well. In particular, the strong scalability of Transformers has allowed them to surpass ConvNets in many areas.

Generative models based on denoising~\citep{ho2020denoising, song2021denoising, dhariwal2021diffusion, rombach2022high, karras2022elucidating, peebles2023scalable, ma2024sit, esser2024scaling} have followed similar architectural trends, ranging from hybrid convolution–attention designs to fully transformer-based backbones.
While foundational works~\citep{ho2020denoising, song2021denoising, dhariwal2021diffusion, rombach2022high, karras2022elucidating} employed a hybrid architecture, DiT~\citep{peebles2023scalable} introduced a fully transformer-based diffusion backbone, replacing convolutions with end-to-end transformer blocks.
This shift has driven the success of recent state-of-the-art diffusion models~\citep{esser2024scaling, flux2024}, offering improved scalability and generation quality. These developments reflect a prevailing belief that scaling transformer-based networks yields better generative performance.
However, such rapid progress has come at the cost of a growing dependence on massive resources, owing to the inherent computational complexity of Transformers. The performance of modern diffusion models is increasingly tied to extensive GPU infrastructure, whose cost and energy demands are becoming major bottlenecks. This trend highlights the need for more efficient diffusion architectures that can achieve competitive performance under practical resource constraints.

\begin{figure}[t]
    \centering
    \includegraphics[width=0.85\linewidth]{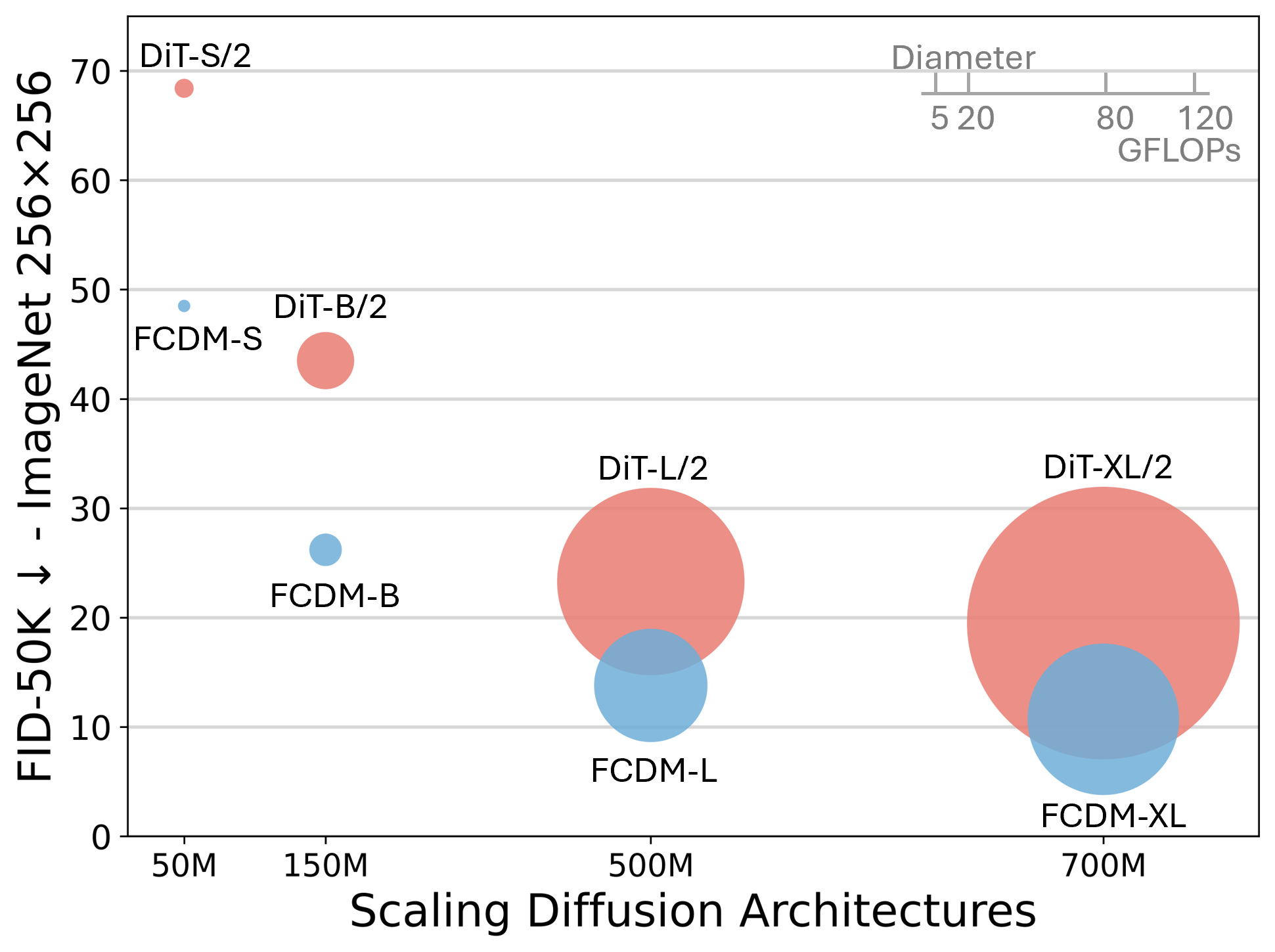}
    \caption{\textbf{Is \emph{scalability} exclusive to transformers?} Our Fully Convolutional Diffusion Model (FCDM) exhibits clear scalability: it is more efficient and achieves better convergence than Diffusion Transformers (DiTs). Bubble size indicates the FLOPs of each diffusion model. Across all scales (ordered by parameter count).}
    \label{fig:fid_flops}
    \vspace{-0.2cm}
\end{figure}

In this work, we revisit the role of convolutions in diffusion modeling and introduce FCDM, a backbone similar to ConvNeXt for generative tasks. 
Building on the architectural strengths of ConvNeXt~\citep{liu2022convnet, woo2023convnext}, which has demonstrated strong competitiveness with Vision Transformers~\citep{dosovitskiy2021an, liu2021swin} in terms of accuracy and scalability on ImageNet classification~\citep{russakovsky2015imagenet}, we design a fully convolutional network for generative diffusion modeling.
Specifically, our architecture differs from ConvNeXt by incorporating conditional injection and organizing it in an easily scalable U-shaped design.
One of the key contributions of DiT~\citep{peebles2023scalable} is its ease of scaling through a small set of intuitive hyperparameters (e.g., number of blocks $L$, hidden channel $C$, number of heads, and patch size $p$), which has made it highly practical and widely adopted in follow-up research.
Our architecture further simplifies this design space, demonstrating an \textit{Easy Scaling Law} with only two hyperparameters (number of blocks $L$ and hidden channel $C$). 
As a fully convolutional architecture, the proposed backbone incorporates the efficiency and scalability of modern ConvNets.

\begin{table}[t]
\centering
\resizebox{\linewidth}{!}{
\begin{tabular}{lcccc}
\toprule[1.5pt]
Model & Train steps & FLOPs (G) $\downarrow$ & Throughput (it/s) $\uparrow$ & FID $\downarrow$ \\
\midrule
DiT-S/2  & 400K & 6   & 1234  & 68.4  \\
FCDM-S   & 400K & \textbf{3}  & \textbf{2687} & \textbf{48.5} \\
\midrule
DiT-B/2  & 400K & 23  & 380.1 & 43.5 \\
FCDM-B   & 400K & \textbf{12} & \textbf{1002} & \textbf{26.2} \\
\midrule
DiT-L/2  & 400K & 81  & 114.6 & 23.3 \\
FCDM-L   & 400K & \textbf{48} & \textbf{381.3} & \textbf{13.8} \\
\midrule
DiT-XL/2 & 400K & 119 & 80.5  & 19.5 \\
FCDM-XL  & 400K & \textbf{65} & \textbf{272.7} & \textbf{10.7} \\
\midrule
DiT-XL/2 & 7M   & 119 & 80.5  & 9.6 \\
FCDM-XL  & 1M   & \textbf{65} & \textbf{272.7} & \textbf{7.9} \\
\bottomrule[1.5pt]
\end{tabular}
}
\caption{FCDM consistently yields lower FLOPs, higher throughput, and converges faster to superior performance compared to DiT across all scales.}
\label{tab:fid_flops}
\vspace{-0.cm}
\end{table}

To isolate architectural effects and ensure a fair comparison of intrinsic network capabilities, we follow the standard DiT training and evaluation framework~\citep{peebles2023scalable}, in line with prior works~\citep{zhu2025dig,tian2025dic,ai2025dico}.
To evaluate scalability, we scale our architecture to match the parameter counts of DiT models and find that our models require approximately 50$\%$ fewer FLOPs.
Furthermore, we observe faster convergence in terms of training steps while achieving competitive performance relative to fully Transformer-based architectures.
As shown in Figure~\ref{fig:fid_flops} and Table~\ref{tab:fid_flops}, our \textbf{F}ully \textbf{C}onvolutional \textbf{D}iffusion \textbf{M}odel (FCDM) is more efficient and converges faster than DiT~\citep{peebles2023scalable} across all model scales.
Through extensive experiments, we find that our architecture remains highly efficient compared to modern diffusion architectures~\citep{zhu2025dig,tian2025dic,ai2025dico}.
These findings re-emphasize the importance of convolutional operations as research increasingly favors Transformer-dominant architectures.
They also offer a complementary perspective for efficiency-focused work: modern fully convolutional architectures provide an alternative path toward scalable, highly efficient generative modeling.
We hope these observations and discussions challenge entrenched assumptions and encourage a reevaluation of the role of convolutions in modern computer vision.

\section{Related Work}
\label{sec:related_works}

This section reviews the architectural evolution of diffusion models, highlighting the transition from convolution--attention hybrid designs to fully transformer-based backbones.  
From these trends, it is evident that fully convolutional diffusion architectures remain largely underexplored compared to their hybrid and transformer counterparts.

\subsection{Hybrid Architectures}
Early diffusion models predominantly adopted hybrid designs, combining convolutional layers for local features with self-attention~\citep{vaswani2017attention} for long-range dependencies.
DDPM~\citep{ho2020denoising}, ScoreSDE~\citep{song2021scorebased}, and DDIM~\citep{song2021denoising} all used hybrid architectures augmented with attention at select resolutions. 
ADM~\citep{dhariwal2021diffusion} further showed that diffusion models could surpass GANs~\citep{goodfellow2014generative, brock2019large, karras2019style} in high-fidelity image generation, solidifying the architecture's viability.
LDM~\citep{rombach2022high} improved scalability by operating in a compressed latent space~\citep{kingma2014auto}, enabling large-scale text-to-image models such as Stable Diffusion and Imagen~\citep{saharia2022photorealistic}.  
Recent works including SDXL~\citep{podell2024sdxl} and SnapGen~\citep{chen2025snapgen} refined training strategies and resolution handling while retaining the convolution–attention hybrid backbone.

\subsection{Transformer-based Diffusion Models}
Transformers~\citep{vaswani2017attention, dosovitskiy2021an} have emerged as strong alternatives, replacing all convolutions with patch-based attention blocks.
DiT~\citep{peebles2023scalable} demonstrated scalability with a ViT-inspired backbone, followed by U-ViT~\citep{bao2023all} and U-DiT~\citep{tian2024u}, which introduced U-shaped variants.
SiT~\citep{ma2024sit} extended DiT to flow matching~\citep{lipman2023flow, liu2023flow}, surpassing DiT across model scales.
PixArt~\citep{chen2024pixartalpha, chen2024pixart}, MM-DiT~\citep{esser2024scaling}, and FLUX~\citep{flux2024} scaled the architecture to production-grade text-to-image models by incorporating improved conditioning pipelines and refined transformer architectures.
Recent approaches such as EQVAE~\citep{kouzelis2025eqvae}, VAVAE~\citep{yao2025reconstruction}, and REPA~\citep{yu2025representation} further accelerated convergence and improved quality.

\subsection{Convolutional Diffusion Models}
Convolutional backbones for diffusion models have only recently reemerged.
DiC~\citep{tian2025dic} re-examined fully convolutional U-Nets using 3$\times$3 convolutions with sparse skip connections, while DiCo~\citep{ai2025dico} adapted 3$\times$3 separable convolutions and proposed compact channel attention to mitigate channel redundancy.
Both methods achieve competitive FID with superior throughput, demonstrating their computational efficiency compared to Diffusion Transformers.
We demonstrate that our revived ConvNeXt backbone not only achieves superior generative performance but also delivers higher computational efficiency than prior convolutional diffusion models, thereby marking a rediscovery of ConvNeXt in the context of efficient generative modeling.

\begin{figure*}[ht]
  \centering
  \vspace{-0.5cm}
    \centerline{{\includegraphics[width=0.87\linewidth]{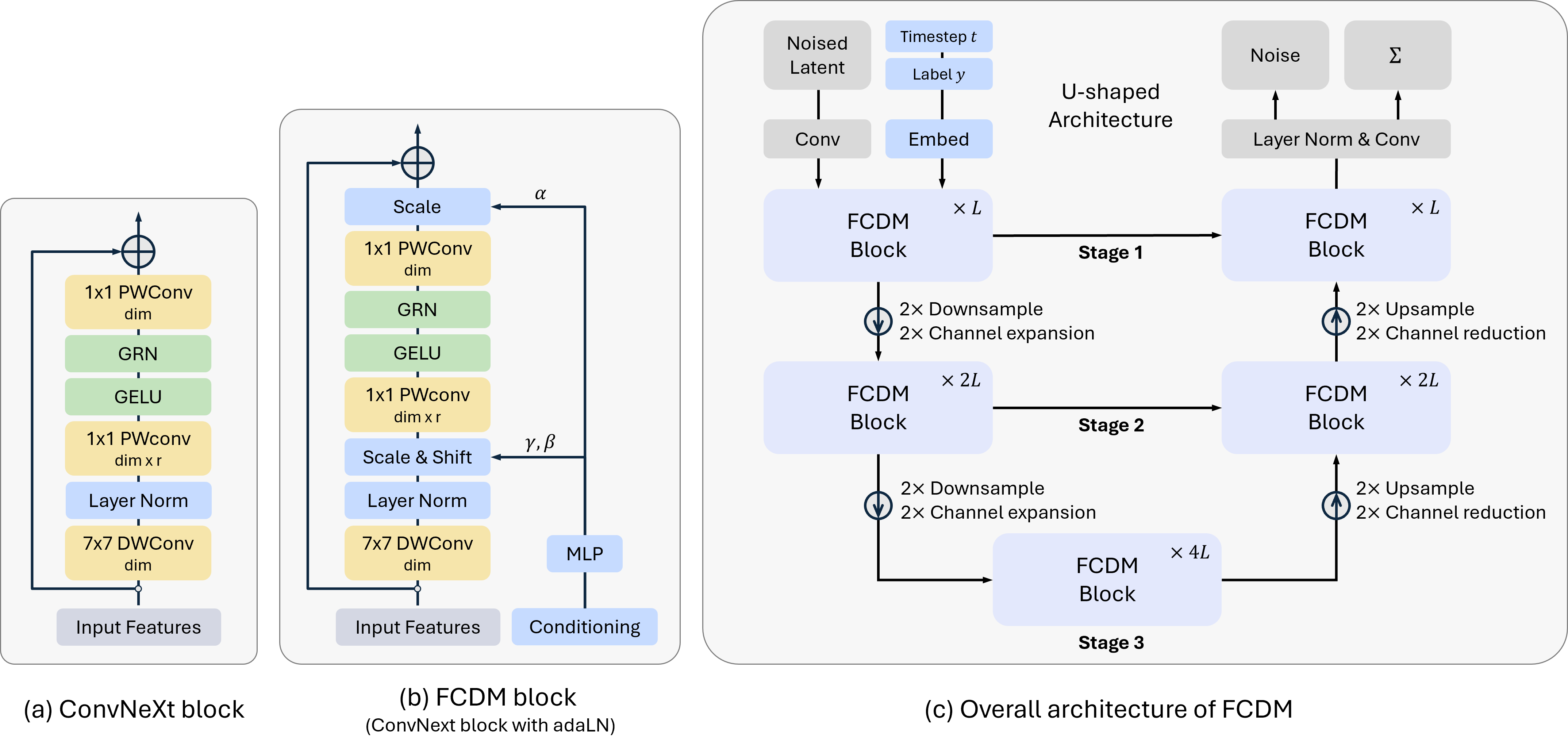}}}
    \caption{
    \textbf{The Fully Convolutional Diffusion Model (FCDM) architecture.}
    (a) Details of the ConvNeXt block.
    (b) Our FCDM block, which incorporates conditioning via adaptive layer normalization.
    (c) We train conditional latent FCDMs. The input latent is processed by multiple FCDM blocks arranged in an easily scalable U-shaped architecture.
    }
    \vspace{-0.3cm}
    \label{fig:architecture}
\end{figure*}

\section{Fully Convolutional Diffusion Models}
\label{sec:method}

We propose a \textbf{F}ully \textbf{C}onvolutional \textbf{D}iffusion \textbf{M}odel (FCDM), reviving the ConvNeXt architecture~\citep{liu2022convnet, woo2023convnext} and adapting it for conditional diffusion generation.
Similar to how DiT~\citep{peebles2023scalable} preserves design practices from Vision Transformers (ViTs)~\citep{dosovitskiy2021an}, FCDM retains the core principles of ConvNeXt. 
While ConvNeXt was originally developed for image classification, diffusion modeling imposes distinct requirements.
We therefore reassemble ConvNeXt with conditional injection, carefully preserving its core design, and make it a suitable backbone for efficient generative diffusion modeling.

\subsection{Designing Diffusion with ConvNeXt}

We redesign ConvNeXt into a generative backbone for diffusion models, introducing the Fully Convolutional Diffusion Model (FCDM).
In this section, we highlight the advantages of our fully convolutional architecture and detail the key components that define the FCDM design space.

\paragraph{ConvNeXt block.}

As shown in Figure~\ref{fig:architecture}~(a), the original ConvNeXt~\citep{liu2022convnet, woo2023convnext} block begins with a 7$\times$7 depthwise convolution, followed by layer normalization~\citep{ba2016layer}.
Two subsequent 1$\times$1 pointwise convolutions handle channel expansion and reduction with a ratio of $r$, while the Global Response Normalization (GRN)~\citep{woo2023convnext} in between mitigates channel redundancy.

\paragraph{Conditional injection.}

The original ConvNeXt blocks lack conditioning mechanisms, as they were originally designed for image classification.
To enable class and time conditioning, we replace LayerNorm with Adaptive LayerNorm (AdaLN), as shown in Figure~\ref{fig:architecture}~(b).
A lightweight MLP maps the conditioning vector (derived from class and time embeddings) to $(\gamma, \beta, \alpha)$ parameters that modulate normalized features.
Following DiT~\citep{peebles2023scalable}, we zero-initialize the final modulation scale $\alpha$ to stabilize optimization and allow deeper training.

\paragraph{Easily scalable U-shaped architecture.}

Most convolutional networks adopt a U-shaped design with skip connections, which facilitates the integration of global and local features.
This structure makes it easier to capture the overall context while preserving the high-resolution details from the early encoder layers.
Following this principle, we organize ConvNeXt blocks within a U-Net hierarchy, with skip connections bridging the encoder and decoder stages. 

To simplify scalability, we avoid the complex, resolution-specific design choices often used in U-shaped networks.
Instead, our architecture is parameterized by only two hyperparameters: the number of blocks $L$ and hidden channels $C$.
At each $2{\times}$ downsampling stage, both $C$ and $L$ are doubled.
This \emph{generalized U-shaped} design (Figure~\ref{fig:architecture}~(c)) allows straightforward scaling while retaining the inductive biases of convolutions.
By controlling $C$ and $L$, the proposed architecture can be scaled up or down flexibly.
Extensive architectural ablations confirm that this simplified design does not compromise performance. For detailed results, please refer to the Supplementary Materials.

\begin{table}[t]
\centering
\resizebox{\linewidth}{!}{%
\begin{tabular}{lccccc}
\toprule
Model (Params) & Blocks $L$ & Channel $C$ & FLOPs (G) & $\frac{\text{FLOPs (FCDM)}}{\text{FLOPs (DiT)}}$ & $\frac{\text{FLOPs (FCDM)}}{\text{FLOPs (DiCo)}}$\\
\midrule
FCDM-S (32.7M)  & 2 & 128 & 3.1 & 50.8$\%$ & 72.9$\%$ \\
FCDM-B (127.7M) & 2 & 256 & 12.2 & 53.0$\%$ & 72.3$\%$ \\
FCDM-L (504.5M)  & 2 & 512 & 48.3 & 59.9$\%$ & 80.2$\%$ \\
FCDM-XL (698.8M) & 3 & 512 & 64.6 & 54.5$\%$ & 74.0$\%$ \\
\bottomrule
\end{tabular}}
\caption{Parameter counts are aligned with the Diffusion Transformer (DiT) configurations for the Small (S), Base (B), Large (L), and XLarge (XL) variants. FLOPs are computed at a 256$\times$256 resolution. Overall, FCDM uses roughly 50$\%$ of the FLOPs of DiT and about 75$\%$ of the FLOPs of DiCo.}
\label{tab:flops_comparison}
\end{table}

\subsection{Revisiting DiCo.}

DiCo~\citep{ai2025dico} represents the current state-of-the-art convolutional diffusion model.
It employs 3$\times$3 separable convolutions and introduces a compact channel attention mechanism that encourages more diverse channel activations.
Interestingly, we observe a close connection between DiCo and our architecture, where the latter offers a more efficient alternative to DiCo’s design choices.
As shown in Table~\ref{tab:flops_comparison}, FCDM achieves approximately 75$\%$ FLOPs efficiency compared to DiCo at a similar parameter scale.
In the following, we detail how our model serves as a more efficient solution for convolutional diffusion models.

\begin{figure}[t]
    \centering
    \vspace{-0.3cm}
    \includegraphics[width=0.85\linewidth]{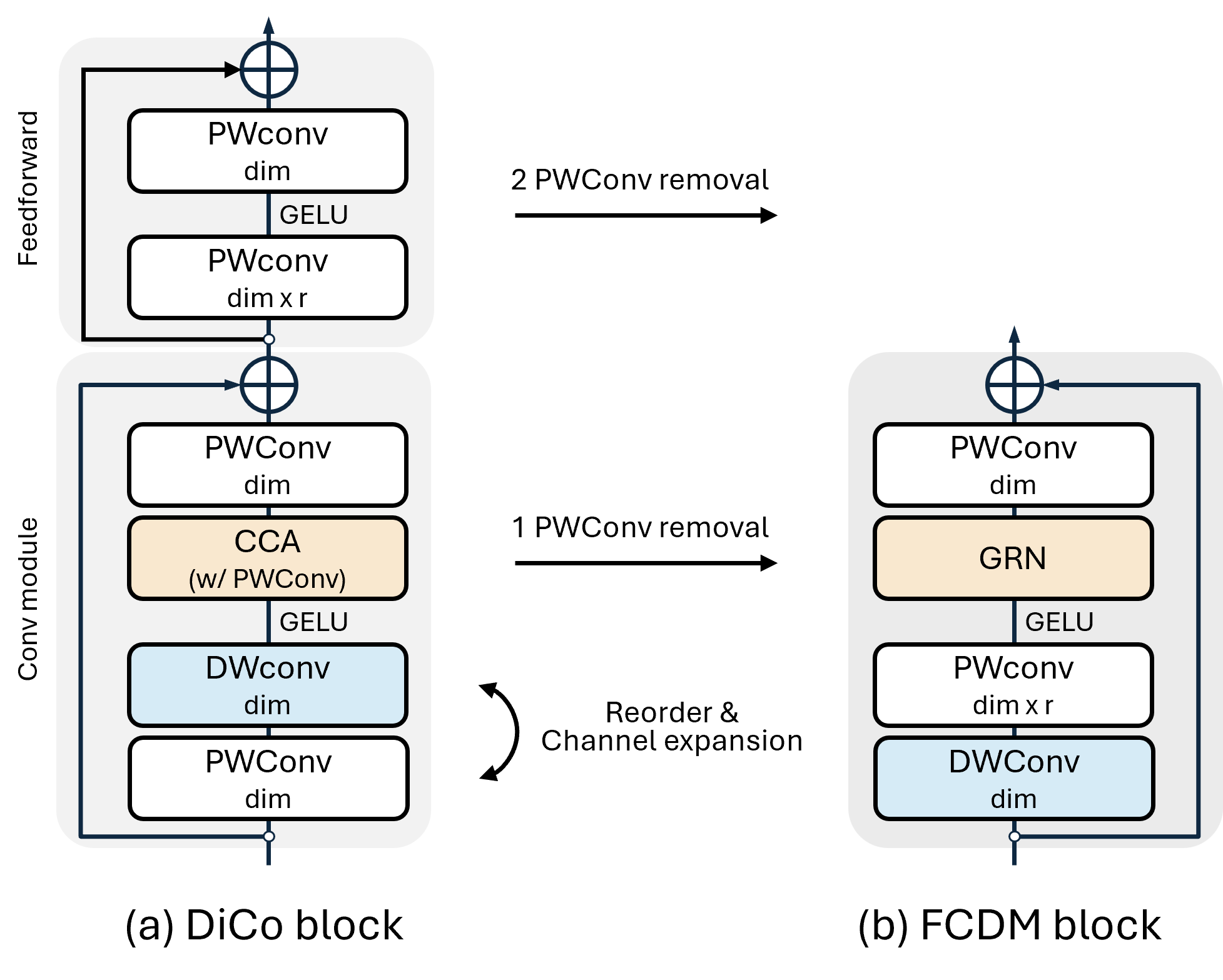}
    \vspace{-0.3cm}
    \caption{\textbf{Simple illustration of DiCo and FCDM block.} Both architectures share a similar high-level structure, but FCDM adopts an inverted bottleneck that expands channels for richer representations while keeping the computational cost of depthwise convolution unchanged. DiCo employs CCA with an additional 1$\times$1 convolution, whereas FCDM uses GRN, requiring no extra pointwise convolutions. FCDM also does not include DiCo’s feed-forward module, resulting in a simpler and more efficient block.}
    \label{fig:block_comparison}
\end{figure}

\paragraph{Richer channel representations.}

As illustrated in Figure~\ref{fig:block_comparison}, the convolution modules of DiCo and FCDM follow a similar high-level structure.  
However, a key structural difference lies in how each model handles channel dimensionality.  
While DiCo preserves the channel dimension throughout the convolution module, our design adopts the inverted bottleneck structure of ConvNeXt, introducing an early channel expansion that allows for richer channel computation within the block.  
Crucially, our design reorders the pointwise and depthwise convolutions so that the channel expansion is applied \textit{after} the depthwise convolution.  
This reordering keeps the computational cost of depthwise convolution unchanged regardless of channel expansion, while still leveraging the expressive capacity of the expanded channels.

\paragraph{GRN promotes diverse channel activations.}

DiCo introduces the compact channel attention (CCA) mechanism to promote diverse channel activations.
We observe that the global response normalization (GRN) layer in ConvNeXt V2~\citep{woo2023convnext} achieves a similar effect while requiring far fewer learnable parameters.
CCA relies on an additional 1$\times$1 pointwise convolution to learn channel-wise activations, whereas GRN is composed primarily of parameter-free operations such as L2 normalization and response normalization~\citep{krizhevsky2012imagenet}.
From this perspective, both mechanisms aim to enhance channel activation diversity, but GRN provides a substantially more efficient alternative.

Lastly, as shown in Figure~\ref{fig:block_comparison}, DiCo includes an additional feed-forward module composed of two 1$\times$1 pointwise convolutions, where the channel expansion is performed.  
In contrast, our design already applies channel expansion within the earlier convolution module through the inverted bottleneck structure.  

Together, these findings indicate that our design provides a more efficient architectural alternative to the current state-of-the-art convolutional diffusion model, DiCo.
In the following section, we show that this improved efficiency is also reflected in the empirical performance.

\section{Experimental Setup}
\label{sec:experimental_setup}

\paragraph{Model size.}

We denote our models by their configurations, parameterized by hidden channels $C$ and number of blocks $L$, which are both doubled at each $2{\times}$ downsampling stage.
To enable fair comparisons, we align the parameter counts of our FCDM scales with those of DiT~\citep{peebles2023scalable} (e.g., DiT-B: 130M vs. FCDM-B: 127.7M).  
We evaluate four model scales, as listed in Table~\ref{tab:flops_comparison}: FCDM-S, FCDM-B, FCDM-L, and FCDM-XL.
These cover a broad range of parameter counts, from 32.7M to 698.8M, allowing us to systematically study scaling behavior and compare with DiT across different scales.

\paragraph{Training.}

We train class-conditional latent FCDMs at $256{\times}256$ and $512{\times}512$ resolutions on the ImageNet dataset~\citep{russakovsky2015imagenet}, a standard yet highly competitive benchmark for generative modeling.
Training follows common practices of DiT~\citep{peebles2023scalable}: we use AdamW~\citep{kingma2015adam, loshchilov2019decoupled} with a fixed learning rate of $1\times 10^{-4}$, no weight decay, and batch size 256.
The only augmentation applied is horizontal flipping.  
We use an exponential moving average (EMA) of model weights with a decay factor of 0.9999, and report all results using the EMA weights.
We retain diffusion hyperparameters from ADM~\citep{dhariwal2021diffusion}: $t_\text{max}=1000$ steps with a linear variance schedule ($1\times 10^{-4}$ to $2\times 10^{-4}$), ADM’s covariance parameterization $\Sigma_\theta$, and their timestep/label embedding method.
Please refer to Supplementary Section~\ref{appendix:ddpm} for an overview of denoising diffusion probabilistic models and to Supplementary Section~\ref{appendix:hyperparameter} for additional training details and hyperparameters.

\paragraph{Datasets and Metrics.}

We conduct experiments on ImageNet-1K at $256{\times}256$ and $512{\times}512$ resolutions for class-conditional image generation.
Our primary metric is Fréchet Inception Distance (FID)~\citep{heusel2017gans}, following the standard evaluation protocol.
We sample 50K images with 250 DDPM sampling steps, and compute the metrics using OpenAI’s official TensorFlow evaluation toolkit~\citep{dhariwal2021diffusion}.
As secondary metrics, we also report Inception Score (IS)~\citep{salimans2016improved} and Precision/Recall~\citep{kynkaanniemi2019improved}.
Please refer to Supplementary Section~\ref{appendix:evaluation} for more details.

\paragraph{Compute.}

All models are implemented in PyTorch~\citep{paszke2019pytorch}.  
The largest model, FCDM-XL, trains at approximately 0.9 iterations per second (with gradient checkpointing) at $256{\times}256$ resolution on four RTX 4090 24GB GPUs with a global batch size of 256, demonstrating the efficiency of our architecture.  
We also verify that the batch size of 256 fits on a single A100 40GB GPU, further highlighting the memory efficiency of our design.

\section{Experiments}
\label{sec:experiments}

\subsection{Scaling model size}

\begin{table*}[t]
\centering
\vspace{-0.5cm}
\resizebox{0.85\textwidth}{!}{
\begin{tabular}{l | c | c c | c c c c}
\toprule
Model & Architecture Type & FLOPs (G) $\downarrow$ & Throughput (it/s) $\uparrow$ & FID $\downarrow$ & IS $\uparrow$ & Precision $\uparrow$ & Recall $\uparrow$ \\
\midrule
DiT-S/2~(400K)~{\scriptsize \textit{(ICCV 2023)}} & Transformer & 6.1 & 1234.0 & 68.40 & - & - & - \\
DiC-S~(400K)~{\scriptsize \textit{(CVPR 2025)}} & Conv & 5.9 & \textbf{3148.8} & 58.68 & 25.82 & - & - \\
DiG-S/2~(400K)~{\scriptsize \textit{(CVPR 2025)}} & Hybrid & 4.3 & 961.2 & 62.06 & 22.81 & 0.39 & 0.56 \\
DiCo-S~(400K)~{\scriptsize \textit{(NeurIPS 2025)}} & Conv & 4.3 & 1695.7 & 49.97 & 31.38 & \textbf{0.48} & \textbf{0.58} \\
\rowcolor{lightgray}
\textbf{FCDM-S}~(400K) & Conv & \textbf{3.1} & 2687.2 & \textbf{48.52} & \textbf{31.64} & \textbf{0.48} & \textbf{0.58} \\
\midrule
DiT-B/2~(400K) & Transformer & 23.0 & 380.1 & 43.47 & - & - & - \\
DiC-B~(400K) & Conv & 23.5 & \textbf{1024.2} & 32.33 & 48.72 & - & - \\
DiG-B/2~(400K) & Hybrid & 17.1 & 345.9 & 39.50 & 37.21 & 0.51 & 
\textbf{0.63} \\
DiCo-B~(400K) & Conv & 16.9 & 823.0 & 27.20 & 56.52 & \textbf{0.60} & 0.61 \\
\rowcolor{lightgray}
\textbf{FCDM-B}~(400K) & Conv & \textbf{12.2} & 1001.6 & \textbf{26.21} & \textbf{58.04} & 0.59 & 0.61 \\
\midrule
DiT-L/2~(400K) & Transformer & 80.7 & 114.6 & 23.33 & - & - & - \\
DiG-L/2~(400K) & Hybrid & 61.7 & 109.0 & 22.90 & 59.87 & 0.60 & \textbf{0.64} \\
DiCo-L~(400K) & Conv  & 60.2 & 288.3 & \textbf{13.66} & \textbf{91.37} & \textbf{0.69} & 0.61 \\
\rowcolor{lightgray}
\textbf{FCDM-L}~(400K) & Conv & \textbf{48.3} & \textbf{381.3} & 13.83 & 93.31 & 0.66 & 0.62 \\
\midrule
DiT-XL/2~(400K) & Transformer & 118.6 & 80.5 & 19.47 & - & - & - \\
DiC-XL~(400K) & Conv & 116.1 & 263.1 & 13.11 & 100.2 & - & - \\
DiG-XL/2~(400K) & Hybrid & 89.4 & 71.7 & 18.53 & 68.53 & 0.63 & \textbf{0.64} \\
DiCo-XL~(400K) & Conv & 87.3 & 174.2 & 11.67 & 100.4 & \textbf{0.71} & 0.61 \\
DiC-H~(400K) & Conv & 204.4 & 144.5 & 11.36 & 106.5 & - & - \\
\rowcolor{lightgray}
\textbf{FCDM-XL}~(400K) & Conv & \textbf{64.6} & \textbf{272.7} & \textbf{10.72} & \textbf{108.0} & 0.69 & 0.63 \\
\midrule
DiT-XL/2~(7M) & Transformer & 118.6 & 80.5 & 9.62 & - & - & - \\
DiC-H~(800K) & Conv & 204.4 & 144.5 & 8.96 & 124.33 & - & - \\
DiG-XL/2~(1.2M) & Hybrid & 89.4 & 71.7 & 8.60 & 130.03 & 0.68 & \textbf{0.68} \\
\rowcolor{lightgray}
\textbf{FCDM-XL}~(1M) & Conv & \textbf{64.6} & \textbf{272.7} & \textbf{7.91} & \textbf{135.55} & \textbf{0.71} & 0.64 \\
\bottomrule
\end{tabular}
}
\caption{\textbf{Scalability comparisons on ImageNet 256$\times$256.} For each model scale, we report FID, IS, Precision, and Recall (50K samples without guidance), and efficiency metrics (training iterations, FLOPs, throughput). FCDM-XL achieves superior convergence while using 50$\%$ fewer FLOPs than DiT-XL/2. The best results are highlighted in \textbf{bold}. Evaluated methods operate in the latent space.}
\label{tab:comparison_scalability}
\end{table*}

\begin{figure*}[t]
  \centering
  \vspace{-0.1cm}
    \centerline{{\includegraphics[width=0.95\linewidth]{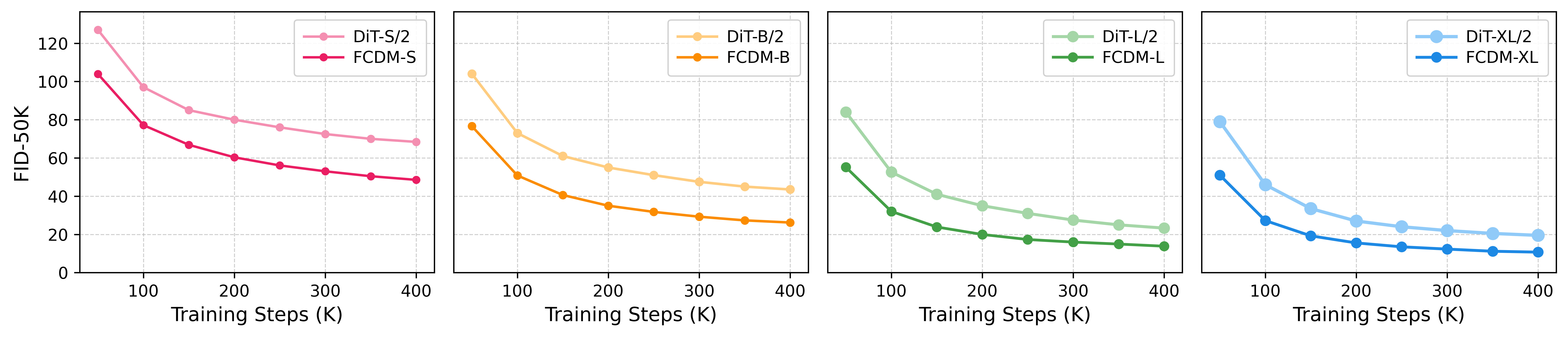}}}
    \vspace{-0.3cm}
    \caption{\textbf{FCDM improves FID across all model scales.} FID-50K over training iterations for both DiT and FCDM. Across all model scales, FCDM converges much faster.}
    \label{fig:scalability}
\end{figure*}

We train four FCDM models (S, B, L, XL), all using the same training configuration.
Figure~\ref{fig:fid_flops} summarizes the FLOPs and FID at 400K training iterations.
In all cases, scaling up model size improves performance.
Figure~\ref{fig:scalability} further shows that FCDM consistently outperforms DiT~\citep{peebles2023scalable} across all scales.
Increasing model scale, both in width and depth, consistently leads to significant FID improvements.

Table~\ref{tab:comparison_scalability} provides a broader comparison with DiT and modern diffusion architectures that follow a similar experimental setup.
Although our scales are aligned with DiT in terms of parameter counts, FCDM requires about 50\% fewer FLOPs than DiT and 25\% fewer FLOPs than DiCo~\citep{ai2025dico}. 
Notably, with 64.6 GFLOPs, FCDM-XL is computationally closer to Large (L) scale models in prior works, yet it outperforms even XL-scale models in terms of FID.
In particular, FCDM-XL achieves superior FID while requiring $7\times$ fewer training steps compared to DiT-XL/2.
Furthermore, this efficiency yields strong throughput performance. While DiC~\citep{tian2025dic} has the best throughput at S and B scales due to its use of standard convolutions, which are simpler and highly optimized for modern hardware, FCDM outperforms it at L and XL scales, achieving the highest throughput.
For completeness, detailed descriptions of each baseline method are provided in Supplementary Section~\ref{appendix:baselines}, and a detailed scaling analysis is provided in Supplementary Section~\ref{appendix:scalability}.

\subsection{Benchmarking Performance and Efficiency}

\begin{table*}[t]
\centering
\vspace{-0.5cm}
\resizebox{0.8\textwidth}{!}{
\begin{tabular}{l c c c c c c c}
\toprule
Model & Training epochs & FLOPs (G) $\downarrow$ & Throughput (it/s) $\uparrow$ & FID $\downarrow$ & IS $\uparrow$ & Precision $\uparrow$ & Recall $\uparrow$ \\
\midrule
\multicolumn{8}{l}{\emph{GAN}} \\
\quad BigGAN-deep & - & - & - & 6.95 & 171.4 & \textbf{0.87} & 0.28 \\
\quad StyleGAN-XL & - & - & - & 2.30 & 265.12 & 0.78 & 0.53 \\
\midrule
\multicolumn{8}{l}{\emph{Pixel diffusion}} \\
\quad ADM-U & 400 & 742.0 & - & 3.94 & 215.8 & 0.83 & 0.53 \\
\quad VDM++ & 560 & - & - & 2.12 & 267.7 & - & - \\
\quad Simple Diffusion & 800 & - & - & 2.77 & 211.8 & - & - \\
\quad CDM & 2160 & - & - & 4.88 & 158.7 & - & - \\
\midrule
\multicolumn{8}{l}{\emph{Latent diffusion}} \\
\quad LDM-4 & 200 & 104.0 & - & 3.60 & 247.7 & \textbf{0.87} & 0.48 \\
\quad U-ViT-H/2 & 240 & 133.3 & 73.5 & 2.29 & 263.9 & 0.82 & 0.57 \\
\quad MaskDiT & 1600 & - & - & 2.28 & 276.6 & 0.80 & \textbf{0.61} \\
\quad SD-DiT & 480 & - & - & 3.23 & - & - & - \\
\midrule
\quad DiT-XL/2 & 1400 & 118.6 & 80.5 & 2.27 & 278.2 & 0.83 & 0.57 \\
\quad SiT-XL/2 & 1400 & 118.6 & 80.5 & 2.06 & 277.5 & 0.83 & 0.59 \\
\quad DiG-XL/2 & 240 & 89.4 & 71.7 & 2.07 & 279.0 & 0.82 & 0.60 \\
\quad DiCo-XL & 750 & 87.3 & 174.2 & 2.05 & 282.2 & 0.83 & 0.59 \\
\quad DiC-H & 400 & 204.4 & 144.5 & 2.25 & - & - & - \\
\rowcolor{lightgray}
\quad \textbf{FCDM-XL} & 400 & \textbf{64.6} & \textbf{272.7} & \textbf{2.03} & \textbf{285.7} & 0.81 & 0.59 \\
\bottomrule
\end{tabular}
}
\vspace{-0.1cm}
\caption{\textbf{Benchmarking class-conditional image generation on ImageNet 256$\times$256.} We compare representative models in terms of FID, IS, Precision, Recall (with guidance), and efficiency metrics (training epochs, FLOPs, throughput). FCDM-XL achieves competitive performance with superior efficiency. The best results are highlighted in \textbf{bold}.}
\label{tab:comparison_256}
\end{table*}

\begin{figure*}[t]
  \centering
  \vspace{-0.1cm}
    \centerline{{\includegraphics[width=0.9\linewidth]{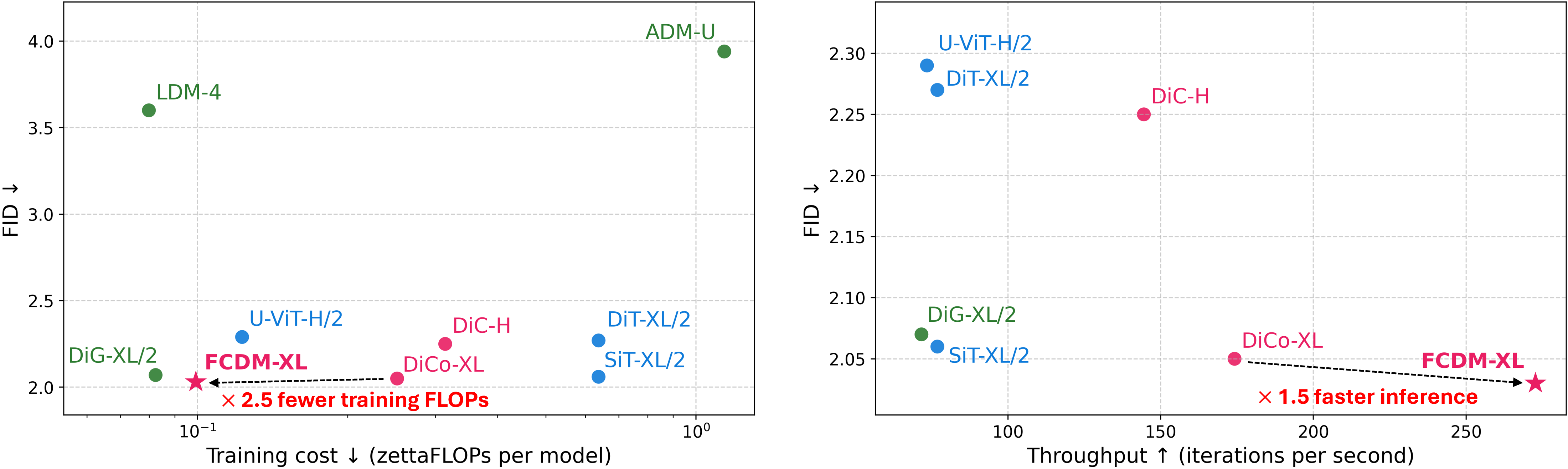}}}
    \caption{\textbf{Benchmarking class-conditional image generation performance and efficiency on ImageNet 256$\times$256.} 
    \textit{Left}: FID versus total training cost. 
    \textit{Right}: FID versus throughput. 
    One zettaFLOP corresponds to $10^{21}$ FLOPs ($10^{12}$ GFLOPs). 
    A training iteration is assumed to cost about $3\times$ one evaluation (forward + backward to inputs + backward to weights). 
    Red denotes fully convolutional, Green denotes hybrid, and Blue denotes fully transformer-based models.
}
    \vspace{-0.2cm}
    \label{fig:training_cost}
\end{figure*}

\paragraph{256$\times$256 ImageNet.}

Building on the scaling analysis, we train FCDM-XL for 2M iterations (400 epochs) and evaluate it with classifier-free guidance~\citep{ho2021classifierfree}.
Figure~\ref{fig:main} presents generated samples and Table~\ref{tab:comparison_256} further compares against prior class-conditional image generation models.
Notably, FCDM-XL improves upon baselines despite requiring fewer training epochs. 
In particular, it achieves an FID of 2.03 and an IS of 285.7, while reaching state-of-the-art efficiency in FLOPs and throughput, demonstrating a strong trade-off between performance and efficiency.
As shown in Figure~\ref{fig:training_cost}, our model achieves state-of-the-art throughput while drastically reducing training cost.
Notably, compared with DiCo, our method maintains competitive performance while operating with 2.5$\times$ fewer total training FLOPs and 1.5$\times$ faster inference throughput.
These results highlight, for the first time, the effectiveness of ConvNeXt architectures in generative diffusion modeling, which had previously been shown only in classification, and demonstrate their broader adaptability and potential.
It is important to note that, although FCDM-XL demonstrates superior performance over comparable baselines within DiT-style experimental settings, it does not yet surpass the latest state-of-the-art results achieved by models such as EDM-2~\citep{karras2024analyzing} or Simpler Diffusion~\citep{hoogeboom2025simpler}. 
Nevertheless, as an architecture with a favorable performance–efficiency trade-off, our approach holds the potential to deliver improved results with further scaling and more advanced training frameworks.

\paragraph{512$\times$512 ImageNet.}

\begin{table*}[t]
\centering
\vspace{-0.5cm}
\resizebox{0.75\textwidth}{!}{
\begin{tabular}{l c c c c c c c}
\toprule
Model & Training iterations & FLOPs (G) $\downarrow$ & Throughput (it/s) $\uparrow$ & FID $\downarrow$ & IS $\uparrow$ & Precision $\uparrow$ & Recall $\uparrow$ \\
\midrule
DiT-XL/2 & 400K & 524.7 & 18.6 & 20.94 & 66.3 & 0.74 & 0.58 \\
DiG-XL/2 & 400K & - & - & 17.36 & 69.4 & 0.75 & \textbf{0.63} \\
DiC-XL & 400K & 464.3 & 124.2 & 15.32 & 93.6 & - & - \\
DiC-H & 400K & 817.2 & 68.6 & 12.89 & 101.8 & - & - \\
\rowcolor{lightgray}
\textbf{FCDM-XL} & 400K & \textbf{257.7} & \textbf{129.6} & \textbf{10.23} & \textbf{108.7} & \textbf{0.79} & 0.60 \\
\midrule
DiT-XL/2 & 1.3M & 524.7 & 18.6 & 13.78 & - & - & - \\
DiT-XL/2 & 3M & 524.7 & 18.6 & 12.03 & 105.25 & 0.75 & \textbf{0.64} \\
DiCo-XL & 1.3M & 349.8 & 82.0 & 8.10 & 132.9 & 0.78 & 0.62 \\
\rowcolor{lightgray}
\textbf{FCDM-XL} & \textbf{1M} & \textbf{257.7} & \textbf{129.6} & \textbf{7.46} & \textbf{133.6} & \textbf{0.79} & 0.61 \\
\bottomrule
\end{tabular}
}
\caption{\textbf{Benchmarking class-conditional image generation on ImageNet 512$\times$512.} 
We report FID, IS, Precision, Recall (without guidance), and efficiency metrics for representative models.
Even at this resolution, FCDM surpasses models trained for 3M iterations with only 1M iterations and achieves the best efficiency in FLOPs and throughput. 
The best results are highlighted in \textbf{bold}.
}
\label{tab:comparison_512}
\end{table*}

\begin{table}[t]
\centering
\resizebox{\columnwidth}{!}{
\begin{tabular}{l c c c}
\toprule
Model & FLOPs (G) $\downarrow$ & FID $\downarrow$ & IS $\uparrow$ \\
\midrule
\textbf{FCDM-L} (Default: 7$\times$7 DWConv) & 48.3 & \textbf{19.97} & \textbf{69.19} \\
\quad $\to$ 5$\times$5 DWConv & 48.2 & 20.48 & 66.69 \\
\quad $\to$ 3$\times$3 DWConv & 48.1 & 21.28 & 64.11 \\
\midrule
\textbf{FCDM-L} (Default: GRN) & 48.3 & \textbf{19.97} & \textbf{69.19} \\
\quad $\to$ CCA$^\ast$~\cite{ai2025dico} & 48.3 & 23.85 & 61.60 \\
\midrule
\textbf{FCDM-L} (Default: w/o Feedforward) & 48.3 & \textbf{19.97} & \textbf{69.19} \\
\quad $\to$ w/ Feedforward$^\ast$ & 48.2 & 28.52 & 47.16 \\
\midrule
\textbf{FCDM-L} (Default: w/ Inv. Bottleneck) & 48.3 & \textbf{19.97} & \textbf{69.19} \\
\quad $\to$ w/o Inv. Bottleneck$^\ast$ & 48.3 & 28.76 & 52.20 \\
\midrule
\textbf{FCDM-L} (Default: FCDM block) & 48.3 & \textbf{19.97} & \textbf{69.19} \\
\quad $\to$ ResNet block$^\ast$ & 48.4 & 31.14 & 49.10 \\
\bottomrule
\end{tabular}
}
\vspace{-0.1cm}
\caption{\textbf{Ablation study on FCDM design choices.} We analyze the effects of kernel size, GRN, DiCo~\citep{ai2025dico} design choices, and the FCDM block. Training iterations are fixed to 200K. $\ast$ indicates that $C$ is adjusted to match FLOPs to ensure a fair comparison.}
\vspace{-0.1cm}
\label{tab:ablation}
\end{table}

\begin{figure*}[t]
  \centering
  \vspace{-0.1cm}
    \centerline{{\includegraphics[width=0.85\linewidth]{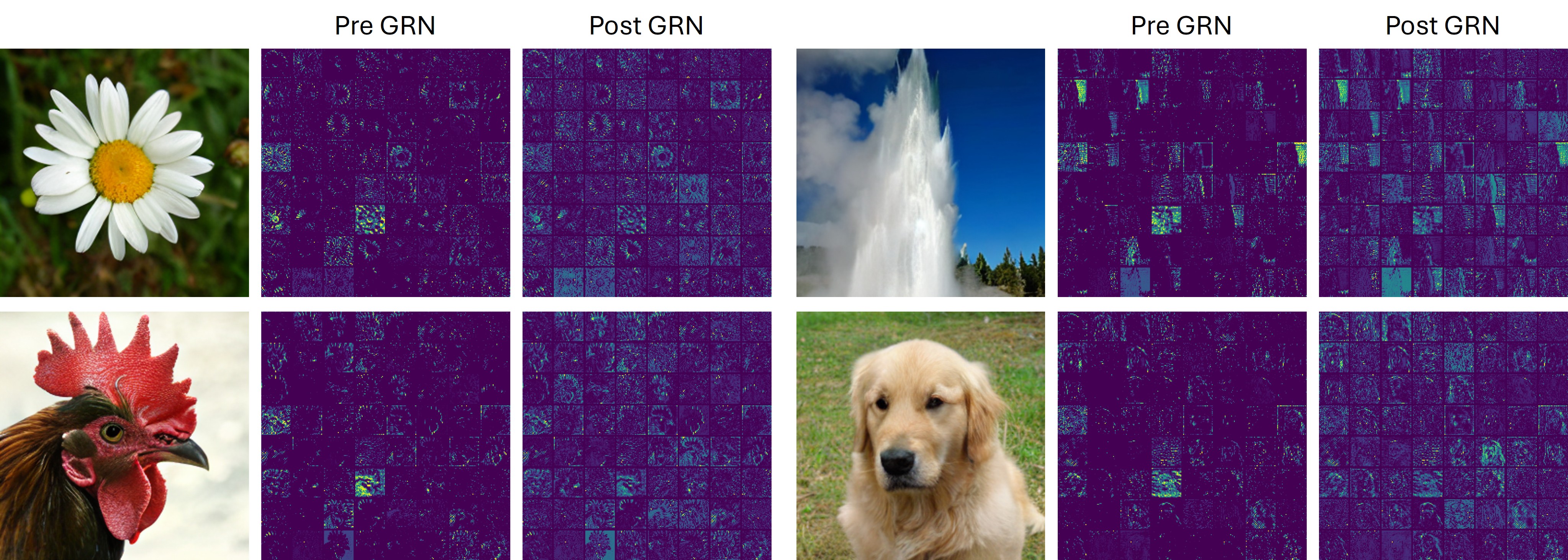}}}
    \vspace{-0.2cm}
    \caption{\textbf{Feature activation visualization.} We visualize features before and after the GRN layer during sampling for each image on the left. The first 64 channels of the last block in the first stage are shown as 8$\times$8 grids. GRN clearly reduces channel redundancy.}
    \label{fig:grn}
\end{figure*}

We also train FCDM-XL at $512{\times}512$ resolution for 1M iterations using the same hyperparameters as the $256{\times}256$ model.
Table~\ref{tab:comparison_512} reports FLOPs and FID at both 400K and 1M iterations.
At this resolution, FCDM-XL again achieves the best FLOPs and throughput.
With this efficiency advantage, FCDM-XL delivers better FID convergence at 400K iterations, and this trend continues at 1M.
Remarkably, FCDM-XL outperforms DiT with 7.5$\times$ fewer training steps.
Moreover, under comparable training iterations, FCDM-XL outperforms DiCo across all evaluation and efficiency metrics.
These results show that FCDM maintains strong efficiency while delivering competitive performance, highlighting the effectiveness of the proposed architecture.
Interestingly, when the resolution doubles, the throughput of DiT drops by approximately 4$\times$, whereas FCDM degrades by only 2$\times$.
This contrast highlights the fundamental computational differences between fully transformer-based and fully convolutional architectures and further illustrates the scalability of FCDM at higher resolutions.
Finally, we provide a frequency-domain analysis in Supplementary Section~\ref{appendix:frequency} to better understand the model behavior compared to Diffusion Transformers, text-to-image results in Supplementary Section~\ref{appendix:text-to-image}, and additional quantitative and qualitative results in Supplementary Sections~\ref{appendix:quantitative} and~\ref{appendix:qualitative}.

\subsection{Ablation Study}

We conduct ablations with the Large (L) model on ImageNet $256{\times}256$ to analyze the effects of key architectural components, as summarized in Table~\ref{tab:ablation}.
First, reducing the kernel size consistently degrades performance, indicating that large kernels successfully expand the effective receptive field to capture broader context.
Second, incorporating DiCo~\citep{ai2025dico} design choices, such as the application of compact channel attention (CCA) or a feedforward module, consistently degrades performance, demonstrating the effectiveness of our design in diffusion modeling.
Finally, removing the inverted bottleneck or replacing FCDM blocks with standard ResNet blocks~\citep{he2016deep} results in a severe degradation, reaffirming the advantage of our design.
Furthermore, as shown in Figure~\ref{fig:grn}, GRN clearly reduces channel redundancy, confirming its contribution to enhancing channel diversity.
These results demonstrate that GRN provides an effect similar to CCA, but with far fewer parameters.
More ablation results are provided in Supplementary Section~\ref{appendix:model_variants}.

\section{Conclusion}
\label{sec:conclusion}

As interest in efficient diffusion architectures continues to grow, we revive the role of ConvNeXt in generative modeling and introduce FCDM, our redesigned ConvNeXt backbone tailored for conditional diffusion models.
By redesigning the ConvNeXt architecture to incorporate conditional injection and organizing it into an easily scalable U-shaped design, we enable FCDM to achieve competitive generative performance with state-of-the-art computational efficiency compared to modern diffusion architectures.
These results demonstrate that modern convolutional architectures provide an alternative path toward scalable, highly efficient generative modeling and challenge the prevailing belief that larger Transformers are the sole path to progress in diffusion models.

\section*{Acknowledgements}
We thank Jakob Buhmann, Farnood Salehi, and Jingwei Tang for helpful discussions.
Taesung Kwon is supported by the National Research Foundation of Korea~(NRF) grant
funded by the Korea government~(MSIT) (RS-2026-25468886).
Taesung Kwon and Jong Chul Ye are further supported by the National Research Foundation of Korea~(NRF)~(RS-2024-00336454 and RS-2023-00262527).
We also acknowledge the support of ETH Z\"{u}rich in providing access to the Euler cluster for this research.

{
    \small
    \bibliographystyle{ieeenat_fullname}
    \bibliography{main}
}

\clearpage
\onecolumn
\begin{center}
    {\Large \textbf{\thetitle}}\\[1.0em]
    {\Large Supplementary Material}\\[2.0em]
\end{center}
\appendix
\setcounter{page}{1}

\noindent We provide the following supplementary sections:
\begin{itemize}
    \setlength{\itemindent}{1em}
    \item Section~\ref{appendix:ddpm}: Overview of denoising diffusion probabilistic models (DDPMs).
    \item Section~\ref{appendix:hyperparameter}: Hyperparameter and implementation details.
    \item Section~\ref{appendix:evaluation}: Descriptions of evaluation metrics.
    \item Section~\ref{appendix:baselines}: Descriptions of baseline models.
    \item Section~\ref{appendix:scalability}: Additional results and analyses on model scalability.
    \item Section~\ref{appendix:frequency}: Frequency-based analysis of model behavior.
    \item Section~\ref{appendix:model_variants}: Additional analyses of architectural variants.
    \item Section~\ref{appendix:text-to-image}: Experimental results on text-to-image generation.
    \item Section~\ref{appendix:quantitative}: Additional quantitative results and analyses.
    \item Section~\ref{appendix:qualitative}: Additional qualitative results and visual samples.
\end{itemize}

\section{Overview of denoising diffusion probabilistic models}
\label{appendix:ddpm}

Diffusion models~\citep{sohl2015deep, ho2020denoising} aim to model a target distribution $p(x)$ by learning a gradual denoising process from Gaussian noise $\mathcal{N}(0,I)$ to $p(x)$.
Specifically, the model learns a \textit{reverse} process $p_\theta(x_{t-1}|x_t)$ of a predefined \textit{forward} diffusion process $q(x_t|x_{t-1})$, which progressively adds Gaussian noise over $T$ timesteps.

For an initial sample $x_0 \sim p(x)$, the \textit{forward} process is defined as:
\begin{equation}
    q(x_t|x_{t-1}) = \mathcal{N}\!\left(x_t; \sqrt{1-\beta_t}\, x_{t-1}, \beta_t I\right),
\end{equation}
where $\beta_t \in (0,1)$ is a variance schedule. A closed-form expression of $q(x_t|x_0)$ can also be derived as:
\begin{equation}
    q(x_t|x_0) = \mathcal{N}\!\left(x_t; \sqrt{\bar{\alpha}_t}\,x_0, (1-\bar{\alpha}_t)I\right), \quad \bar{\alpha}_t = \prod_{s=1}^t (1-\beta_s).
\end{equation}

The denoising diffusion probabilistic model (DDPM)~\citep{ho2020denoising} parameterizes the \textit{reverse} transition as:
\begin{equation}
    p_\theta(x_{t-1}|x_t) =
    \mathcal{N}\Bigl(x_{t-1};
    \frac{1}{\sqrt{\alpha_t}}\Bigl(x_t -
    \frac{1-\alpha_t}{\sqrt{1-\bar{\alpha}_t}}\, \epsilon_\theta(x_t,t)\Bigr),
    \sigma_t^2 I\Bigr),
\end{equation}
where the noise predictor $\epsilon_\theta(x_t,t)$ is trained using the simple denoising autoencoder objective:
\begin{equation}
    \mathcal{L}_{\text{simple}} =
    \mathbb{E}_{x_t, x_0,\epsilon,t}\left[\left\|
    \epsilon - \epsilon_\theta(x_t,t)\right\|^2_2\right].
\end{equation}

Following DDPM, one can set $\sigma_t^2 = \beta_t$ for simplicity.
Meanwhile, improved DDPM (iDDPM)~\citep{nichol2021improved} shows that performance can be improved by jointly learning the variance $\Sigma_\theta(x_t, t)$, which is parameterized as an interpolation between $\beta_t$ and $\tilde{\beta}_t$ in the log domain:
\begin{equation}
\log \Sigma_\theta(x_t, t) = v \log \beta_t + (1 - v) \, \log \tilde{\beta}_t,
\end{equation}
where $\tilde{\beta}_t = \frac{1-\bar{\alpha}_{t-1}}{1-\bar{\alpha}_{t}}\beta_t$, and $v$ denotes the interpolation weight predicted by the model in a dimension-wise manner.
In this work, we adopt the iDDPM framework for both training and sampling, following the same design choice as DiT~\citep{peebles2023scalable}.

\clearpage

\section{Hyperparameters and implementation details}
\label{appendix:hyperparameter}

We design Fully Convolutional Diffusion Models (FCDMs) at multiple scales, aligned by parameter counts with DiT.
Thanks to their easy scalability, we adjust only two hyperparameters, $L$ and $C$, to obtain these variants.
Notably, when compared to DiT in terms of FLOPs, our FCDMs require only 50.8$\%$ to 59.9$\%$ of those consumed by DiT, demonstrating the state-of-the-art computational efficiency of our design.
Our implementation is based on the original DiT codebase~\citep{peebles2023scalable}.
Detailed configurations of these hyperparameters, along with additional implementation details, are provided in Table~\ref{tab:appendix_hyperparameter}.
For the latent space, we adopt an off-the-shelf pre-trained variational autoencoder (VAE)~\citep{kingma2014auto, rombach2022high, kouzelis2025eqvae} with a downsampling factor of 8.
Accordingly, an input RGB image of shape $256{\times}256{\times}3$ is encoded to a latent tensor of $32{\times}32{\times}4$.
All diffusion training operates in this latent space, and latents are decoded back to pixels by the VAE decoder.

\begin{table*}[ht]
\centering
\resizebox{0.75\textwidth}{!}{
\begin{tabular}{lccccc}
\toprule
 & FCDM-S & FCDM-B & FCDM-L & FCDM-XL & FCDM-XL \\
Resolution & 256$\times$256 & 256$\times$256 & 256$\times$256 & 256$\times$256 & 512$\times$512 \\
\midrule 
\textbf{Architecture} & & & & & \\
Input dim. & 32$\times$32$\times$4 & 32$\times$32$\times$4 & 32$\times$32$\times$4 & 32$\times$32$\times$4 & 64$\times$64$\times$4 \\
Num. blocks ($L$) & 2 & 2 & 2 & 3 & 3 \\
Hidden channels ($C$) & 128 & 256 & 512 & 512 & 512 \\
1$\times$1 conv. expansion ratio ($r$) & 3 & 3 & 3 & 3 & 3 \\
\midrule
\textbf{Training} & & & & & \\
Training iteration & 400K & 400K & 400K & 2M & 1M \\
Global batch size & 256 & 256 & 256 & 256 & 256 \\
Optimizer & AdamW & AdamW & AdamW & AdamW & AdamW \\
Learning rate & $1 \times 10^{-4}$ & $1 \times 10^{-4}$ & $1 \times 10^{-4}$ & $1 \times 10^{-4}$ & $1 \times 10^{-4}$ \\
Learning rate schedule & constant & constant & constant & constant & constant \\
($\beta_1, \beta_2$) & (0.9, 0.999) & (0.9, 0.999) & (0.9, 0.999) & (0.9, 0.999) & (0.9, 0.999) \\
Weight decay & 0 & 0 & 0 & 0 & 0 \\
Numerical precision & fp32 & fp32 & fp32 & fp32 & fp32 \\
Data augmentation & random flip & random flip & random flip & random flip & random flip\\
\midrule
\textbf{Sampling} & & & & & \\
Sampler & iDDPM & iDDPM & iDDPM & iDDPM & iDDPM \\
Sampling steps & 250 & 250 & 250 & 250 & 250 \\
\bottomrule
\end{tabular}}
\caption{\textbf{Hyperparameter setup of FCDM model scales.} For all scales of FCDM, we adopt the same experimental settings as DiT.}
\label{tab:appendix_hyperparameter}
\end{table*}

\paragraph{Computing resources.}
Thanks to the superior efficiency of our fully convolutional architecture, we are able to train ImageNet models at 256$\times$256 resolution using \textit{consumer}-grade GPUs such as the NVIDIA RTX 4090 (24GB).
For 256$\times$256, we train FCDM-XL with a batch size of 256 on 4$\times$ RTX 4090 GPUs, achieving a training throughput of approximately 0.9 steps/s with gradient checkpointing enabled.
We also confirm that the same batch size 256 fits on a single NVIDIA A100 40GB GPU, further demonstrating the memory efficiency of our design.
For 512$\times$512, we use 4$\times$ NVIDIA H100 80GB GPUs and obtain a throughput of about 0.7 steps/s (also with gradient checkpointing) with the same batch size.

\clearpage

\section{Evaluation metrics}
\label{appendix:evaluation}

For evaluation, we follow the setup of ADM~\citep{dhariwal2021diffusion} and use the same reference batches provided in their official implementation.
Specifically, we generate 50K samples and compute the metrics using OpenAI's official TensorFlow evaluation toolkit.
All evaluations are conducted on NVIDIA RTX 4090 or NVIDIA H100 GPUs, except for certain reported numbers that are taken from prior work.

The following gives a concise description of the evaluation metrics used in our experiments.

\begin{itemize}
    \item \textbf{FID}~\citep{heusel2017gans} measures the distance between the feature distributions of real and generated images. It is computed using the Inception-v3 network~\citep{szegedy2016rethinking}, under the assumption that both feature distributions follow multivariate Gaussian distributions.
    \item \textbf{sFID}~\citep{nash2021generating} computes FID using spatial feature maps from intermediate layers of Inception-v3, thereby better capturing the spatial structure of generated images.
    \item \textbf{IS}~\citep{salimans2016improved} evaluates only generated images using the Inception-v3 network. It assigns higher scores when the images are classifiable with high confidence (sharp and meaningful) and when the set of generated images is diverse across different categories.
    \item \textbf{Precision and Recall}~\citep{kynkaanniemi2019improved} measure realism and diversity in feature space. Precision reflects the fraction of generated images that look realistic, while recall reflects how much of the real data distribution is covered by the generated samples.
\end{itemize}

We additionally report computational efficiency.
FLOPs are computed using {\small \texttt{torchprofile}}, and throughput is evaluated under the sampling configurations of DiT~\citep{peebles2023scalable} with a batch size of 64.
FlashAttention-2~\citep{dao2024flashattention} and Flash Linear Attention~\citep{yang2024gated} are activated in DiT and DiG, respectively.

\section{Baseline models}
\label{appendix:baselines}

The following summarizes the key ideas of the diffusion baselines used for the evaluation.

\begin{itemize}
    \item \textbf{ADM}~\citep{dhariwal2021diffusion} improves hybrid U-Net architecture for diffusion models and introduces classifier guidance, which enables a trade-off between sample quality and diversity.
    \item \textbf{VDM++}~\citep{kingma2023understanding} enhances training efficiency by proposing a simple adaptive noise schedule for diffusion models.
    \item \textbf{Simple diffusion}~\citep{hoogeboom2023simple} proposes a diffusion model for high-resolution image generation by carefully redesigning the noise schedule and model architecture.
    \item \textbf{CDM}~\citep{ho2022cascaded} adopts a cascaded framework in which a base model first generates a low-resolution image, and subsequent super-resolution diffusion models progressively refine it to higher fidelity.
    \item \textbf{LDM}~\citep{rombach2022high} proposes latent diffusion models that operate in a compressed latent space, greatly improving training efficiency while retaining high generation quality.
    \item \textbf{U-ViT}~\citep{bao2023all} adapts Vision Transformers for latent diffusion by introducing long skip connections similar to those in U-Net.
    \item \textbf{MaskDiT}~\citep{zheng2024fast} proposes an asymmetric encoder–decoder architecture for diffusion transformers, trained with an auxiliary mask reconstruction task to improve efficiency.
    \item \textbf{SD-DiT}~\citep{zhu2024sd} reframes the mask modeling of MaskDiT as a self-supervised discrimination objective.
    \item \textbf{DiT}~\citep{peebles2023scalable} replaces the hybrid U-Net architecture with a fully transformer-based backbone, introduces AdaIN-zero conditioning to stabilize training, and shows that diffusion transformers scale effectively.
    \item \textbf{SiT}~\citep{ma2024sit} reformulates DiT training by transitioning from discrete diffusion to continuous flow matching.
    \item \textbf{DiG}~\citep{zhu2025dig} integrates Gated Linear Attention~\citep{yang2024gated}, enabling sub-quadratic complexity efficiency of diffusion transformers.
    \item \textbf{DiC}~\citep{tian2025dic} re-examines purely convolutional denoisers by scaling standard 3$\times$3 convolutional blocks in a U-Net design, introducing sparse skip connections.
    \item \textbf{DiCo}~\citep{ai2025dico} proposes a 3$\times$3 separable convolutional block in a U-Net design, introducing compact channel attention to activate more informative channels.
\end{itemize}

\newpage

\section{Additional Scaling Results}
\label{appendix:scalability}

As shown in Figure~\ref{fig:scalability}, we clearly demonstrate the scalability of FCDM in terms of FID.
We also observe consistent scalability across other metrics, including sFID, Inception Score, Precision, and Recall, as reported in Table~\ref{tab:scalability_diffusion}.

In addition, we trained FCDMs using the original SiT implementation~\citep{ma2024sit} to examine whether our proposed design also exhibits scalability under this framework.
Following the original implementation details, we trained the model with the flow-matching objective~\citep{lipman2023flow,ma2024sit}.
We used AdamW with a constant learning rate of $1\times10^{-4}$, $(\beta_1,\beta_2)=(0.9,0.999)$, and no weight decay.
For sampling, we employed the Euler--Maruyama SDE sampler with 250 steps, setting the final step size to 0.04.

As shown in Table~\ref{tab:scalability_flow}, we again observe clear scalability when training the proposed network with the flow-matching objective.
Interestingly, under our framework, the flow-matching objective yields better performance at the Small (S) scale, while showing slightly worse results at the Base (B) through XLarge (XL) scales.
Nevertheless, these results confirm that the proposed architecture possesses generalized scalability beyond a specific training objective.

\begin{table*}[ht]
\centering
\resizebox{0.75\textwidth}{!}{
\begin{tabular}{l|c|c|ccc|cc}
\toprule
Model & FLOPs (G) & Training Steps & FID $\downarrow$ & sFID $\downarrow$ & IS $\uparrow$ & Precision $\uparrow$ & Recall $\uparrow$ \\
\midrule
\multirow{8}{*}{FCDM-S} & \multirow{8}{*}{3.10}
& 50K  & 103.93 & 15.03 & 12.01 & 0.3013 & 0.3513 \\
& & 100K & 77.17  & 12.15 & 17.11 & 0.3809 & 0.4473 \\
& & 150K & 66.85  & 11.02 & 20.68 & 0.4136 & 0.5106 \\
& & 200K & 60.33  & 10.70 & 23.71 & 0.4396 & 0.5537 \\
& & 250K & 56.08  & 10.54 & 26.30 & 0.4568 & 0.5609 \\
& & 300K & 53.01  & 10.59 & 28.10 & 0.4687 & 0.5806 \\
& & 350K & 50.44  & 10.29 & 30.08 & 0.4757 & 0.5729 \\
& & 400K & \textbf{48.53}  & \textbf{10.12} & \textbf{31.64} & \textbf{0.4836} & \textbf{0.5840} \\
\midrule
\multirow{8}{*}{FCDM-B} & \multirow{8}{*}{12.20}
& 50K  & 76.61 & 9.75 & 16.45 & 0.3797 & 0.4569 \\
& & 100K & 50.83 & 8.24 & 26.90 & 0.4854 & 0.5543 \\
& & 150K & 40.60 & 7.53 & 35.35 & 0.5268 & 0.5894 \\
& & 200K & 35.00 & 7.22 & 42.42 & 0.5525 & 0.5981 \\
& & 250K & 31.77 & 7.11 & 47.22 & 0.5665 & 0.6117 \\
& & 300K & 29.26 & 7.03 & 51.62 & 0.5705 & 0.6141 \\
& & 350K & 27.34 & 6.94 & 55.10 & 0.5855 & 0.6127 \\
& & 400K & \textbf{26.21} & \textbf{6.86} & \textbf{58.04} & \textbf{0.5908} & \textbf{0.6112} \\
\midrule
\multirow{8}{*}{FCDM-L} & \multirow{8}{*}{48.30}
& 50K  & 55.15 & 8.33 & 22.74 & 0.4655 & 0.5403 \\
& & 100K & 32.03 & 7.10 & 43.20 & 0.5791 & 0.5857 \\
& & 150K & 23.88 & 6.47 & 58.51 & 0.6106 & 0.6052 \\
& & 200K & 19.97 & 6.17 & 69.19 & 0.6312 & 0.6128 \\
& & 250K & 17.33 & 5.99 & 77.79 & 0.6427 & 0.6233 \\
& & 300K & 15.98 & 5.82 & 84.28 & 0.6477 & 0.6245 \\
& & 350K & 14.95 & 5.82 & 87.55 & 0.6527 & 0.6257 \\
& & 400K & \textbf{13.83} & \textbf{5.65} & \textbf{93.31} & \textbf{0.6612} & \textbf{0.6218} \\
\midrule
\multirow{8}{*}{FCDM-XL} & \multirow{8}{*}{64.60}
& 50K  & 51.00 & 8.31 & 24.37 & 0.4940 & 0.5475 \\
& & 100K & 27.23 & 6.86 & 49.52 & 0.6108 & 0.5824 \\
& & 150K & 19.25 & 6.15 & 68.62 & 0.6492 & 0.5995 \\
& & 200K & 15.54 & 5.95 & 81.40 & 0.6690 & 0.6051 \\
& & 250K & 13.50 & 5.74 & 91.74 & 0.6785 & 0.6117 \\
& & 300K & 12.31 & 5.64 & 98.58 & 0.6829 & 0.6192 \\
& & 350K & 11.19 & 5.54 & 104.86 & 0.6914 & 0.6227 \\
& & 400K & \textbf{10.72} & \textbf{5.47} & \textbf{108.04} & \textbf{0.6864} & \textbf{0.6273} \\
\bottomrule
\end{tabular}
}
\caption{\textbf{Performance of FCDMs across scales and training steps on ImageNet 256$\times$256 (Diffusion).} 
Scaling FCDMs consistently leads to improved generative performance when trained with the diffusion objective.}
\label{tab:scalability_diffusion}
\end{table*}

\begin{table*}[ht]
\centering
\resizebox{0.75\textwidth}{!}{
\begin{tabular}{l|c|c|ccc|cc}
\toprule
Model & FLOPs (G) & Training Steps & FID $\downarrow$ & sFID $\downarrow$ & IS $\uparrow$ & Precision $\uparrow$ & Recall $\uparrow$ \\
\midrule
\multirow{8}{*}{FCDM-S} & \multirow{8}{*}{3.10}
& 50K  & 103.10 & 13.10 & 12.23 & 0.2863 & 0.3111 \\
& & 100K & 76.95 & 11.12 & 16.96 & 0.3826 & 0.4547 \\
& & 150K & 66.97 & 10.18 & 20.56 & 0.4228 & 0.4953 \\
& & 200K & 60.53 & 9.62 & 23.90 & 0.4428 & 0.5372 \\
& & 250K & 55.94 & 9.53 & 26.16 & 0.4670 & 0.5515 \\
& & 300K & 52.43 & 9.22 & 28.78 & 0.4787 & 0.5632 \\
& & 350K & 49.76 & 9.05 & 31.06 & 0.4866 & 0.5730 \\
& & 400K & \textbf{47.84} & \textbf{8.91} & \textbf{32.89} & \textbf{0.4944} & \textbf{0.5749} \\
\midrule
\multirow{8}{*}{FCDM-B} & \multirow{8}{*}{12.20}
& 50K  & 80.10 & 18.48 & 15.46 & 0.3286 & 0.4072 \\
& & 100K & 52.23 & 8.34 & 25.95 & 0.4891 & 0.5450 \\
& & 150K & 42.58 & 7.85 & 33.83 & 0.5346 & 0.5780 \\
& & 200K & 37.01 & 7.51 & 40.72 & 0.5554 & 0.5819 \\
& & 250K & 33.14 & 7.24 & 46.39 & 0.5728 & 0.5855 \\
& & 300K & 30.16 & 7.07 & 51.60 & 0.5883 & 0.5953 \\
& & 350K & 28.40 & 6.99 & 55.07 & 0.5941 & 0.6066 \\
& & 400K & \textbf{26.61} & \textbf{6.85} & \textbf{58.51} & \textbf{0.6050} & \textbf{0.6017} \\
\midrule
\multirow{8}{*}{FCDM-L} & \multirow{8}{*}{48.30}
& 50K  & 57.32 & 15.82 & 21.61 & 0.4467 & 0.4872 \\
& & 100K & 33.71 & 7.74 & 40.77 & 0.5852 & 0.5615 \\
& & 150K & 26.10 & 6.94 & 54.79 & 0.6232 & 0.5865 \\
& & 200K & 21.91 & 6.63 & 65.12 & 0.6429 & 0.5909 \\
& & 250K & 19.39 & 6.47 & 73.55 & 0.6557 & 0.5940 \\
& & 300K & 17.59 & 6.31 & 80.47 & 0.6650 & 0.6075 \\
& & 350K & 16.27 & 6.18 & 85.62 & 0.6681 & 0.6064 \\
& & 400K & \textbf{15.30} & \textbf{6.17} & \textbf{90.09} & \textbf{0.6751} & \textbf{0.6126} \\
\midrule
\multirow{8}{*}{FCDM-XL} & \multirow{8}{*}{64.60}
& 50K  & 51.21 & 11.89 & 24.21 & 0.5010 & 0.5006 \\
& & 100K & 29.37 & 7.34 & 46.27 & 0.6172 & 0.5680 \\
& & 150K & 22.07 & 6.77 & 63.02 & 0.6572 & 0.5870 \\
& & 200K & 17.98 & 6.34 & 75.71 & 0.6747 & 0.5909 \\
& & 250K & 15.72 & 6.24 & 85.30 & 0.6853 & 0.5991 \\
& & 300K & 14.16 & 6.07 & 92.79 & 0.6952 & 0.6040 \\
& & 350K & 13.06 & 5.97 & 98.22 & 0.6976 & 0.6072 \\
& & 400K & \textbf{12.11} & \textbf{5.96} & \textbf{103.07} & \textbf{0.7030} & \textbf{0.6070} \\
\bottomrule
\end{tabular}
}
\caption{\textbf{Performance of FCDMs across scales and training steps on ImageNet 256$\times$256 (Flow-Matching).} 
Scaling FCDMs also demonstrates consistent improvements in generative performance when trained with the flow-matching objective.}
\label{tab:scalability_flow}
\end{table*}

\clearpage

\section{Frequency-based analysis}
\label{appendix:frequency}

\begin{figure}[ht]
    \centering
    \includegraphics[width=0.8\linewidth]{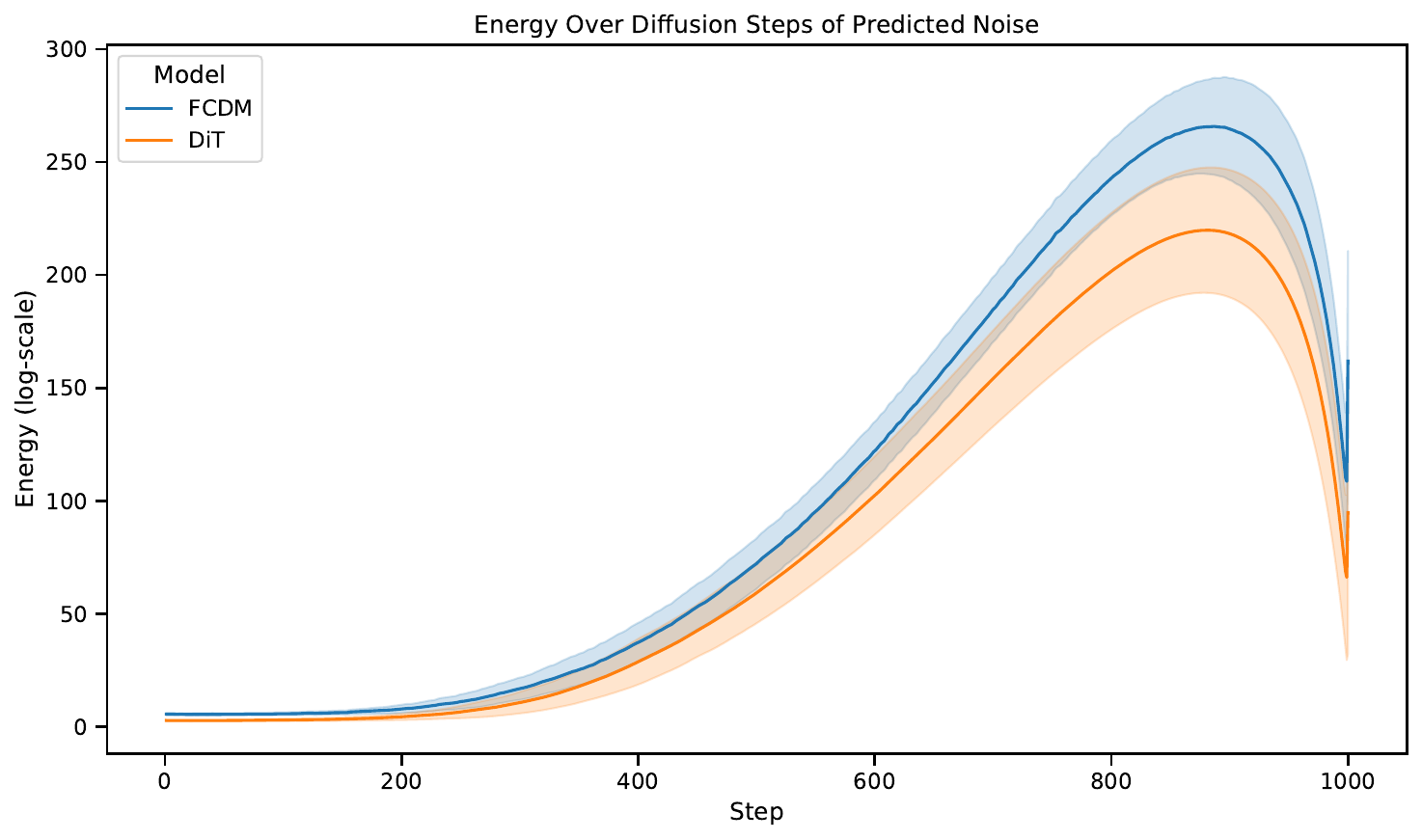}
    \caption{\textbf{Spectral energy of predicted noise across diffusion steps.} FCDM consistently exhibits higher spectral energy than DiT across the entire diffusion process, suggesting potential for better preservation of high-frequency components.}
    \label{fig:energy_plot}
\end{figure}

To better highlight the differences between our fully convolutional diffusion model (FCDM) and the transformer-based DiT, we examine the evolution of the spectral energy, defined as the sum of the log-magnitude spectrum of the predicted noise, over the course of the diffusion process (using models trained for 400K iterations at 
512$\times$512 resolution).
For each predicted noise sample, we compute the 2D Fourier transform, take the magnitude spectrum, and apply a logarithmic scaling $\log(1+F)$ to compress the dynamic range.
We then define the total spectral energy as the sum of all values in this log-magnitude spectrum, which reflects the overall distribution of frequency components.
Figure~\ref{fig:energy_plot} presents the total spectral energy, averaged over 128 validation samples, calculated at each of the 1,000 diffusion timesteps.

Across all diffusion steps, FCDM consistently exhibits higher spectral energy than DiT.
This difference is most pronounced in the early-to-middle stages of the diffusion trajectory, where the model must simultaneously capture global structure and fine-grained detail.
The elevated energy of FCDM indicates that its predicted noise retains stronger high-frequency components, which can be associated with sharper textures, edges, and local structures.
By contrast, DiT produces lower spectral energy, suggesting smoother predictions with fewer high-frequency details.
While this observation may provide a partial explanation for the performance gap between FCDM and DiT, further theoretical analysis is required.

\clearpage

\section{Analysis of additional architectural variants}
\label{appendix:model_variants}

This section provides further details on the ablations in the main manuscript and introduces additional architectural ablation results.
Table~\ref{tab:ablation} presents architectural ablations using the Large (L) model on ImageNet at $256{\times}256$.
We analyze the effects of specific architectural elements, including kernel size, Global Response Normalization (GRN), the inverted bottleneck (channel expansion), the feedforward module, and FCDM blocks. 
In addition, we evaluate various other design choices to further investigate their impact on performance and efficiency.
Interestingly, although ConvNeXt~\citep{liu2022convnet, woo2023convnext} was developed for a different task, we observe similar ablation trends in our experiments.

\paragraph{Effect of Kernel Size.}

Vision Transformers employ non-local self-attention, enabling each layer to access a global receptive field.
In contrast, ConvNets traditionally rely on stacking small 3$\times$3 convolutions (popularized by VGGNet~\citep{simonyan2015very}), which are efficient on modern GPUs~\citep{lavin2016fast}.
Our experiments show that reducing the kernel size from the default 7$\times$7 consistently degrades performance, with FID increasing from 19.97 to 20.48 and 21.28.
This suggests that larger kernels provide a larger effective receptive field, allowing the model to capture broader context.

Additionally, we explore kernel sizes larger than the default 7$\times$7.
Interestingly, as shown in Table~\ref{tab:ablation_kernel}, performance degrades for kernel sizes beyond 7$\times$7, which may slow down convergence speed and leave the models under-trained within our 200K iteration ablation studies.

Taken together, these results suggest that the 7$\times$7 kernel provides the best balance between expressiveness and convergence speed for our architecture.

\begin{table}[h]
\centering
\resizebox{0.48\textwidth}{!}{
\begin{tabular}{l c c c c}
\toprule
Model Configuration & FLOPs (G) $\downarrow$ & FID $\downarrow$ & IS $\uparrow$ \\
\midrule
\textbf{FCDM-L} (Default: 7$\times$7 DWConv) & 48.3 & \textbf{19.97} & \textbf{69.19} \\
\quad 7$\times$7 $\to$ 9$\times$9 DWConv & 48.4 & 23.85 & 59.68 \\
\quad 7$\times$7 $\to$ 11$\times$11 DWConv & 48.6 & 23.93 & 60.18 \\
\bottomrule
\end{tabular}
}
\caption{\textbf{Ablation study on larger kernels.} We analyze effects of larger kernel sizes. Training iterations are fixed to 200K.}
\label{tab:ablation_kernel}
\end{table}

\paragraph{Effect of Applying DiCo Design Choices.}

Following DiCo~\citep{ai2025dico}, we investigate whether the architectural design choices from DiCo improve performance in our framework.
For a fair comparison, we adjust the channel dimensions ($C$) to match the total FLOPs across variants.

First, we replace the GRN layer with CCA~\citep{ai2025dico}, as both modules are designed to enhance channel diversity.
As shown in Table~\ref{tab:ablation}, using CCA degrades performance compared to GRN, indicating that GRN is a more effective channel enhancement module in our architecture.
Second, we remove channel expansion to align with DiCo, which does not apply channel expansion in its convolutional blocks.
As shown in Table~\ref{tab:ablation}, disabling channel expansion leads to a clear drop in performance, demonstrating that channel expansion is a crucial component of our architecture for maintaining generative capacity.
Lastly, we add the feedforward module after the convolutional block to follow the DiCo design, whereas our baseline does not include this component.
As shown in Table~\ref{tab:ablation}, adding the feedforward module further degrades performance, suggesting that our design is better suited for efficient diffusion modeling.

Overall, these results indicate that our design choices consistently lead to superior performance at matched FLOPs, validating the effectiveness of our architecture.

\paragraph{Effect of FCDM Blocks.}

We replaced FCDM blocks with ResNet blocks~\citep{he2016deep} using standard $3{\times}3$ convolutions.
To match FLOPs, the hidden channels were reduced from 512 to 336, given the higher computational cost of standard convolutions compared to separable convolutions.
This substitution results in a substantial degradation, with FID increasing from 19.97 to 31.14, indicating that the FCDM block is better suited for this task than the ResNet block.

\paragraph{Effect of Autoencoders.}

Since FCDM operates in latent space, we tested whether performance persists under different VAEs.
As shown in Table~\ref{tab:ablation_autoencoder}, FCDM consistently outperforms DiT under both SD-VAE~\citep{rombach2022high} and EQ-VAE~\citep{kouzelis2025eqvae}.
Similar to DiT, our model performs best with EQ-VAE, improving further over SD-VAE.  
These findings suggest that techniques originally proposed to enhance DiT (e.g., stronger VAEs) transfer equally well to FCDM, indicating the potential for further performance improvements.

\begin{table*}[h]
\centering
\resizebox{0.48\textwidth}{!}{
\begin{tabular}{l c c c c c c}
\toprule
Model Configuration & Training iterations & FLOPs (G) $\downarrow$ & FID $\downarrow$ \\
\midrule
\multicolumn{4}{l}{\emph{SD-VAE}} \\
\quad DiT-XL/2 & 400K & 118.6 & 19.47 \\
\quad FCDM-XL & 400K & \textbf{64.6} & \textbf{11.57} \\
\midrule
\multicolumn{4}{l}{\emph{EQ-VAE}} \\
\quad DiT-XL/2 & 400K & 118.6 & 14.50 \\
\quad FCDM-XL & 400K & \textbf{64.6} & \textbf{10.72} \\
\bottomrule
\end{tabular}
}
\caption{\textbf{Ablation study on autoencoders.} Across different latent spaces (SD-VAE and EQ-VAE), FCDM consistently outperforms DiT.}
\label{tab:ablation_autoencoder}
\end{table*}

\paragraph{Effect of Isotropic Architecture.}

To further analyze our architecture, we evaluated an isotropic variant of our model. As shown in Table~\ref{tab:isotropic}, despite using only $\sim$48\% of the parameters of DiT-B/2~\citep{peebles2023scalable}, our isotropic variant achieves a lower FID given the same number of training steps. This again validates the effectiveness and efficiency of our fully convolutional block design.

\begin{table}[ht]
\centering
\caption{\textbf{Ablation study on isotropic architecture.} Comparison of our isotropic variant against DiT-B/2~\citep{peebles2023scalable}. All models are trained on ImageNet $256 \times 256$ for 200k iterations under identical settings.
}
\label{tab:isotropic}
\resizebox{0.33\linewidth}{!}{
\begin{tabular}{l c c}
\toprule
Model Configuration & Params (M) & FID $\downarrow$ \\
\midrule
DiT-B/2 & 130 & 55.00 \\
\textbf{FCDM (Iso.)} & \textbf{62 ($\sim$48\%)} & \textbf{41.15} \\
\bottomrule
\end{tabular}
}
\end{table}

\paragraph{Effect of Replacing Convolution with Local Attention.}

We conduct an ablation study to examine the impact of replacing depthwise convolution with local self-attention in our architecture.
Specifically, we employ Neighborhood Attention (NA)~\citep{hassani2023neighborhood}, which implements sliding-window local attention with efficient C++ and CUDA kernels.
For a fair comparison, we use a 7$\times$7 attention window to match the receptive field of our 7$\times$7 depthwise convolution.
We also adjust the channel dimensions to ensure matched FLOPs across variants.

As shown in Table~\ref{tab:ablation_nat}, replacing depthwise convolution with NA results in a significant performance drop in terms of both FID and IS.
Moreover, the throughput is substantially reduced, indicating that local self-attention is considerably less efficient in practice.

These results suggest that depthwise convolution is more effective than local self-attention in terms of both performance and efficiency in our architecture, making it a better choice for efficient diffusion modeling.

\begin{table}[h]
\centering
\resizebox{0.55\textwidth}{!}{
\begin{tabular}{l c c c c}
\toprule
Model Configuration & FLOPs (G) $\downarrow$ & TP (it/s) $\uparrow$ & FID $\downarrow$ & IS $\uparrow$ \\
\midrule
\textbf{FCDM-L} (Default: DWConv) & 48.3 & \textbf{381.3} & \textbf{19.97} & \textbf{69.19} \\
\quad DWConv $\to$ NA$^\ast$~\citep{hassani2023neighborhood} & 45.9 & 122.8 & 29.81 & 50.92 \\
\bottomrule
\end{tabular}
}
\caption{\textbf{Ablation study on replacing depthwise convolution with local self-attention.}
Neighborhood Attention (NA)~\citep{hassani2023neighborhood} with a 7$\times$7 window is used as a replacement for depthwise convolution.
All models are trained for 200K iterations at matched FLOPs. $\ast$ indicates that $C$ is adjusted to match FLOPs to ensure a fair comparison.}
\label{tab:ablation_nat}
\end{table}

\paragraph{Effect of Asymmetric Encoder–Decoder Allocation.}

Following~\citep{hoogeboom2025simpler}, we investigated whether an asymmetric allocation of compute between the encoder and decoder could outperform the symmetric setup.
Intuitively, assigning more compute to the decoder appears advantageous, since upsampling from low to high resolution is more demanding and additionally requires processing skip connections.
As shown in Table~\ref{tab:model_comparison}, an asymmetric design slightly improves FID for the Large (L) model (19.97 $\to$ 19.55).
However, for the XLarge (XL) model, the asymmetric setup performs on par with the symmetric variant.
While asymmetric encoder–decoder architectures remain an interesting direction, we adopt the symmetric setup for its simplicity and more straightforward scalability.

\begin{table*}[h]
\centering
\caption{\textbf{Ablation study on asymmetric block allocation.} All models trained for 200K iterations under identical training settings.
}
\label{tab:model_comparison}
\resizebox{0.9\textwidth}{!}{
\begin{tabular}{l c c c c c c}
\toprule
Model Configuration & Hidden channel $C$ & Depths & Params (M) & FLOPs (G) $\downarrow$ & TP (it/s) $\uparrow$ & FID $\downarrow$ \\
\midrule
\multicolumn{7}{l}{\textit{Asymmetric U-Net Ablations}} \\
\midrule
\quad \textbf{FCDM-L} (Default: Sym. U-Net) & 512 & {[2, 4, 8, 4, 2]} & 504.5 & 48.3 & 381.3 & 19.97 \\
\quad \quad Asym. U-Net & 512 & {[2, 3, 8, 5, 2]} & 504.5 &48.3 & 381.3 & \textbf{19.55} \\
\midrule
\quad \textbf{FCDM-XL} (Default: Sym. U-Net) & 512 & {[3, 6, 12, 6, 3]} & 698.8 & 64.6 & 272.7 & \textbf{15.54} \\
\quad \quad Asym. U-Net & 512 & {[3, 5, 12, 7, 3]} & 698.8 & 64.6 & 272.7 & \textbf{15.54} \\
\quad \quad Asym. U-Net & 512 & {[3, 3, 12, 9, 3]} & 698.8 & 64.6 & 272.7 & 15.55 \\
\bottomrule
\end{tabular}
}
\end{table*}

\paragraph{Effect of Block Scaling across U-Net Levels.}

We investigated how block scaling across U-Net levels affects model behavior. We compared (1) a default scaling setup, where deeper levels have more blocks, and (2) a uniform setup, where all levels use the same number of blocks. To ensure fairness, we kept the total number of blocks fixed and matched the total parameter count by adjusting channel sizes.
As shown in Table~\ref{tab:scaling_laws}, the uniform setup yields better FID for the Large (L) model (19.97 $\to$ 17.63). However, it significantly degrades efficiency: FLOPs increase from 48.3G to 62.8G, reaching a level similar to the XLarge (XL) model with the scaling setup (62.8G vs. 64.6G), while still yielding worse FID (17.63 vs. 15.54).
Given this trade-off, we adopt the default scaling setup in our architecture, since efficiency is a core design goal of this work.

\begin{table}[ht]
\centering
\caption{\textbf{Ablation study on block scaling strategies.} All models are trained on ImageNet $256 \times 256$ for 200k iterations under identical settings. The channel size is adjusted to match the total number of parameters.
}
\label{tab:scaling_laws}
\resizebox{0.85\linewidth}{!}{
\begin{tabular}{l c c c c c c}
\toprule
Model Configuration & Hidden channel $C$ & Depths & Params (M) & FLOPs (G) $\downarrow$ & TP (it/s) $\uparrow$ & FID $\downarrow$ \\
\midrule
\textbf{FCDM-L} (Default: Scaling) & 512 & {[2, 4, 8, 4, 2]} & 504.5 & \textbf{48.3} & \textbf{381.3} & 19.97 \\
\quad Uniform & 600 & {[4, 4, 4, 4, 4]} & 496.6 & 62.8 & 261.6 & 17.63 \\
\midrule
\textbf{FCDM-XL} (Default: Scaling) & 512 & {[3, 6, 12, 6, 3]} & 698.8 & 64.6 & 272.7 & \textbf{15.54} \\
\bottomrule
\end{tabular}
}
\end{table}

\clearpage

\section{Text-to-image generation experiment}
\label{appendix:text-to-image}

\begin{figure}[h]
  \centering
    \centerline{{\includegraphics[width=0.9\linewidth]{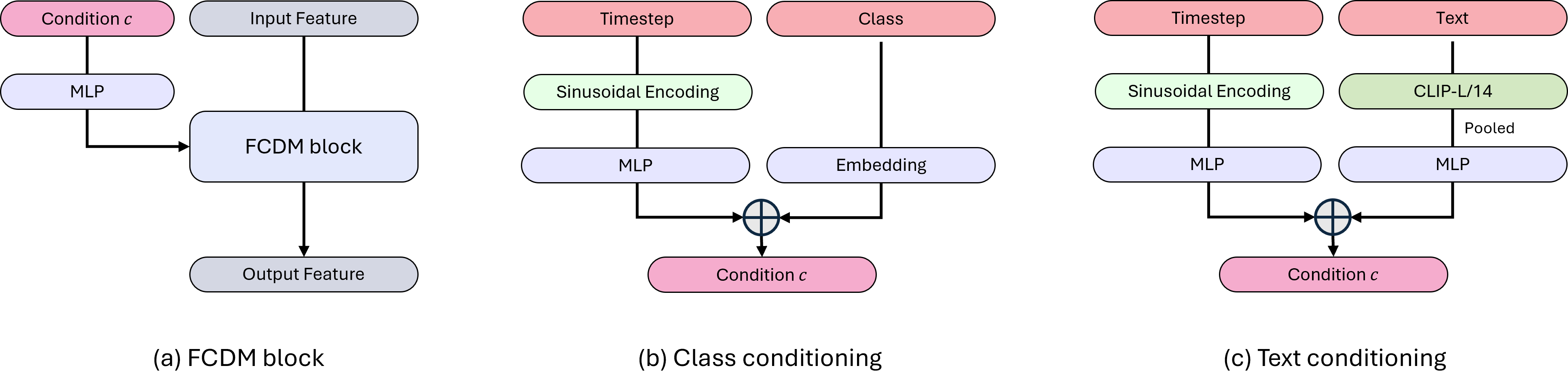}}}
    \vspace{-0.3cm}
    \caption{
    \textbf{Conditioning modules for class and text in the FCDM architecture.}
    (a) FCDM block with conditioning vector $c$,
    (b) Conditioning module for class conditioning,
    (c) Conditioning module for text conditioning incorporating the CLIP text encoder.
    }
    \label{fig:text_conditioning}
\end{figure}

Originally, the FCDM block (Figure~\ref{fig:text_conditioning}~(a)) is designed for class-conditioned generation, where the conditioning vector $c$ is derived from the diffusion timestep and the class label using the conditioning module illustrated in Figure~\ref{fig:text_conditioning}~(b).
To adapt FCDM for text-to-image generation, we modify this conditioning module by replacing the class embedding layer with a CLIP text encoder~\citep{radford2021learning} followed by MLP layers, as shown in Figure~\ref{fig:text_conditioning}~(c).
This design enables the network to be conditioned on both the diffusion timestep and the pooled text representation~\citep{podell2024sdxl, esser2024scaling}.
By modifying only the conditioning module while preserving the rest of the architecture, we can reuse the same FCDM backbone for text-to-image generation.

We train FCDM-XL with the text-conditioning module from scratch on the MS-COCO~\citep{lin2014microsoft} training split and evaluate it on the validation split.
The model is trained for 100K iterations with a batch size of 256, using the CLIP text encoder to compute pooled text embeddings.
Apart from the modified conditioning module, all other training hyperparameters and settings are kept identical to those of the original FCDM configuration.

\begin{figure}[h]
  \centering
    \centerline{{\includegraphics[width=0.9\linewidth]{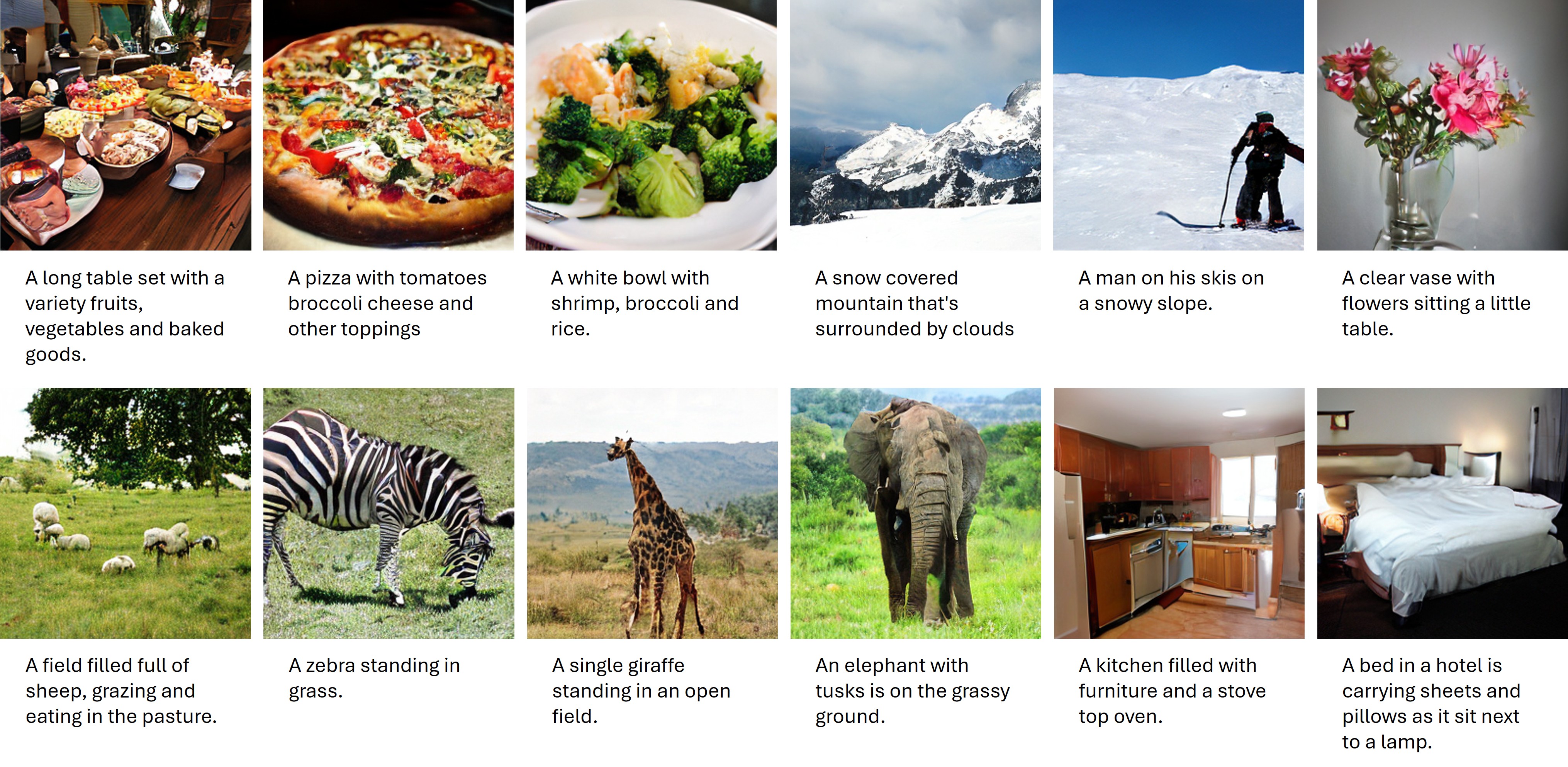}}}
    \vspace{-0.3cm}
    \caption{
    \textbf{Qualitative results on text-to-image generation (MS-COCO).} We use classifier-free guidance with scale = 4.0.
    }
    \label{fig:ms_coco}
\end{figure}

As shown in Figure~\ref{fig:ms_coco}, FCDM successfully generates images corresponding to the given text descriptions, even when trained for only 100K iterations using pooled text embeddings alone.
These findings indicate that text conditioning can be effectively incorporated into FCDM.
Nevertheless, extending FCDM to support joint conditioning on full text embeddings, as in MMDiT~\citep{esser2024scaling}, represents an important direction for future work toward learning richer representations and improving generation performance.

\clearpage

\section{More quantitative results}
\label{appendix:quantitative}

While we position our architecture as a state-of-the-art model in terms of \textit{efficiency}, it is also important to assess how its \textit{performance} relates to other diffusion models.
To this end, building upon the results in Table~\ref{tab:comparison_scalability} and Table~\ref{tab:comparison_256}, we extend our evaluation to additional baselines~\citep{gao2023masked, yao2024fasterdit}, focusing on both efficiency and performance.

As presented in Table~\ref{tab:appendix_scalability} and Table~\ref{tab:appendix_comparison_256}, our findings indicate that FCDM-XL achieves generation quality closely aligned with FasterDiT~\citep{yao2024fasterdit}.
Despite this similarity in performance, our architecture provides substantial efficiency benefits: it achieves 3.3$\times$ faster inference throughput and requires 1.8$\times$ fewer total training FLOPs.

These results reaffirm that our architecture remains highly competitive, offering competitive performance while maintaining state-of-the-art efficiency.
We hope that our architecture will serve as a practical and efficient backbone for future research.

\begin{table*}[h]
\centering
\resizebox{0.9\textwidth}{!}{
\begin{tabular}{l | c | c c | c c c c}
\toprule
Model & Architecture Type & FLOPs (G) $\downarrow$ & Throughput (it/s) $\uparrow$ & FID $\downarrow$ & IS $\uparrow$ & Precision $\uparrow$ & Recall $\uparrow$ \\
\midrule
MDT-XL/2~(400K)~\citep{gao2023masked} & Transformer & 118.7 & 83.9 & 16.42 & - & - & - \\
FasterDiT-XL/2~(400K)~\citep{yao2024fasterdit} & Transformer & 118.6 & 80.5 & 11.90 & - & - & - \\
\rowcolor{lightgray}
\textbf{FCDM-XL}~(400K) & Conv & \textbf{64.6} & \textbf{272.7} & \textbf{10.72} & 108.0 & 0.69 & 0.63 \\
\midrule
MDT-XL/2~(1.3M) & Transformer & 118.7 & 83.9 & 9.60 & - & - & - \\
FasterDiT-XL/2~(1M) & Transformer & 118.6 & 80.5 & 8.72 & 121.17 & 0.68 & \textbf{0.67} \\
\rowcolor{lightgray}
\textbf{FCDM-XL}~(1M) & Conv & \textbf{64.6} & \textbf{272.7} & \textbf{7.91} & \textbf{135.55} & \textbf{0.71} & 0.64 \\
\bottomrule
\end{tabular}
}
\caption{\textbf{Additional comparisons on ImageNet 256$\times$256 without guidance.}
We report efficiency metrics (FLOPs, throughput) and performance metrics (FID, IS, Precision, Recall) using 50K samples. The best results are shown in \textbf{bold}.}
\label{tab:appendix_scalability}
\end{table*}

\begin{table*}[h]
\centering
\resizebox{\textwidth}{!}{
\begin{tabular}{l c c c c c c c c}
\toprule
Model & Training epochs & FLOPs (G) $\downarrow$ & Training FLOPs (Z) $\downarrow$ & Throughput (it/s) $\uparrow$ & FID $\downarrow$ & IS $\uparrow$ & Precision $\uparrow$ & Recall $\uparrow$ \\
\midrule
MDT-XL/2~\citep{gao2023masked} & 500 & 118.7 & 0.228 & 83.9 & 2.15 & 249.3 & \textbf{0.82} & 0.58 \\
FasterDiT-XL/2~\citep{yao2024fasterdit} & 400 & 118.6 & 0.182 & 80.5 & \textbf{2.03} & 270.0 & 0.81 & \textbf{0.60} \\
\rowcolor{lightgray}
\textbf{FCDM-XL} & 400 & \textbf{64.6} & \textbf{0.099} & \textbf{272.7} & \textbf{2.03} & \textbf{285.7} & 0.81 & 0.59 \\
\bottomrule
\end{tabular}
}
\caption{\textbf{Additional comparisons on ImageNet 256$\times$256 with guidance.}
We report efficiency metrics (FLOPs, training FLOPs, throughput) and performance metrics (FID, IS, Precision, Recall) using 50K samples. The best results are shown in \textbf{bold}.}
\label{tab:appendix_comparison_256}
\end{table*}

\clearpage

\section{More qualitative results}
\label{appendix:qualitative}

\begin{figure}[ht]
  \centering
  \begin{subfigure}[t]{0.49\columnwidth}
    \centering
    \includegraphics[width=\linewidth]{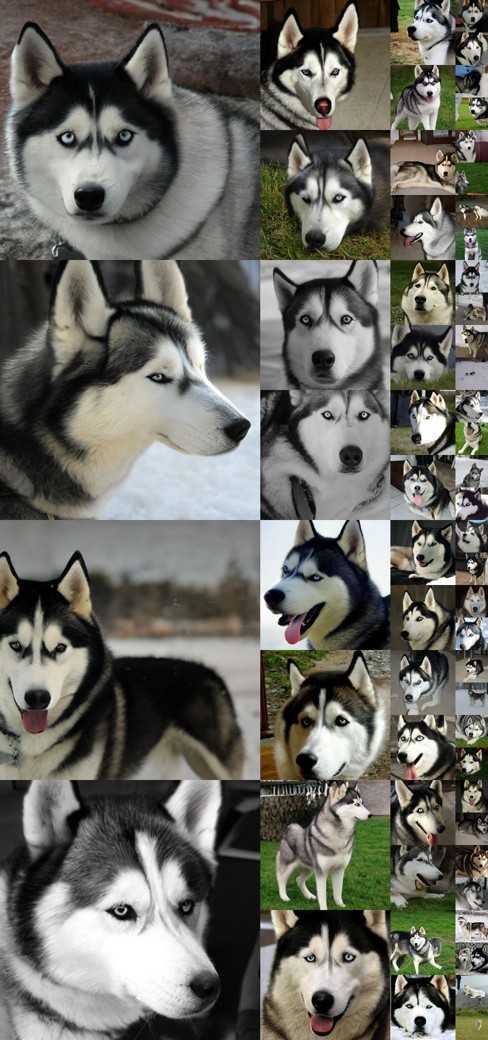}
    \caption*{Class label = ``husky'' (250)}
  \end{subfigure}\hfill
  \begin{subfigure}[t]{0.49\columnwidth}
    \centering
    \includegraphics[width=\linewidth]{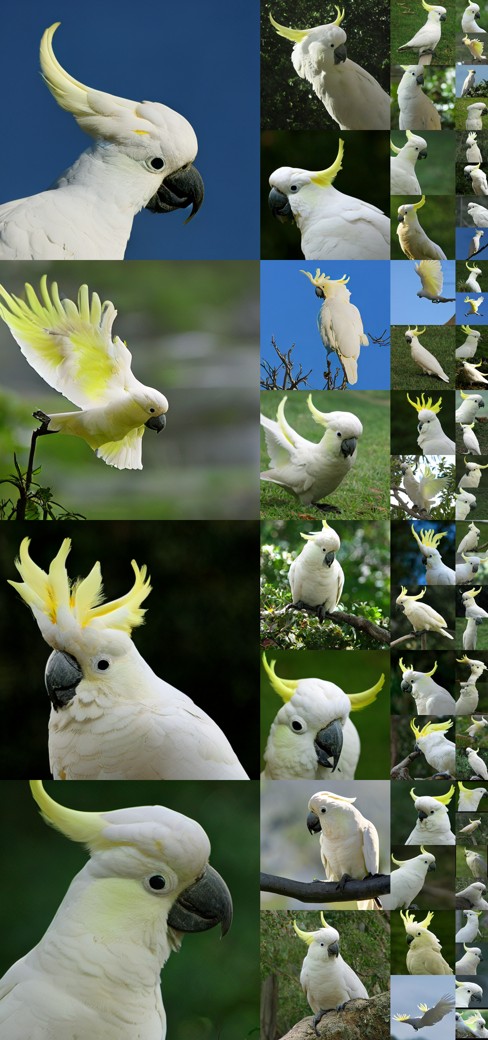}
    \caption*{Class label = ``sulphur-crested cockatoo'' (89)}
  \end{subfigure}
  \caption{\textbf{Uncurated 512$\times$512 FCDM-XL samples.} Classifier-free guidance scale = 4.0}
  \label{appendix_fig1}
\end{figure}

\begin{figure}[ht]
  \centering
  \begin{subfigure}[t]{0.49\columnwidth}
    \centering
    \includegraphics[width=\linewidth]{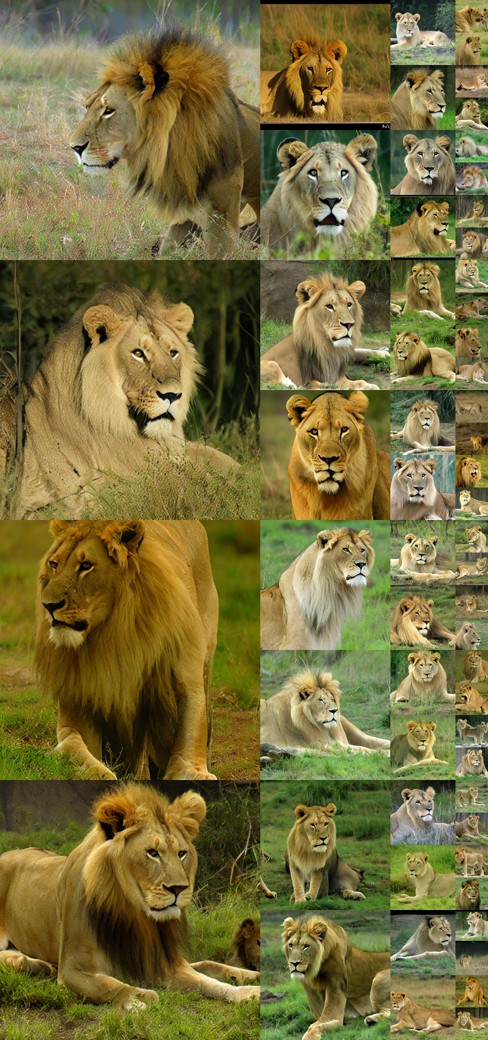}
    \caption*{Class label = ``lion'' (291)}
  \end{subfigure}\hfill
  \begin{subfigure}[t]{0.49\columnwidth}
    \centering
    \includegraphics[width=\linewidth]{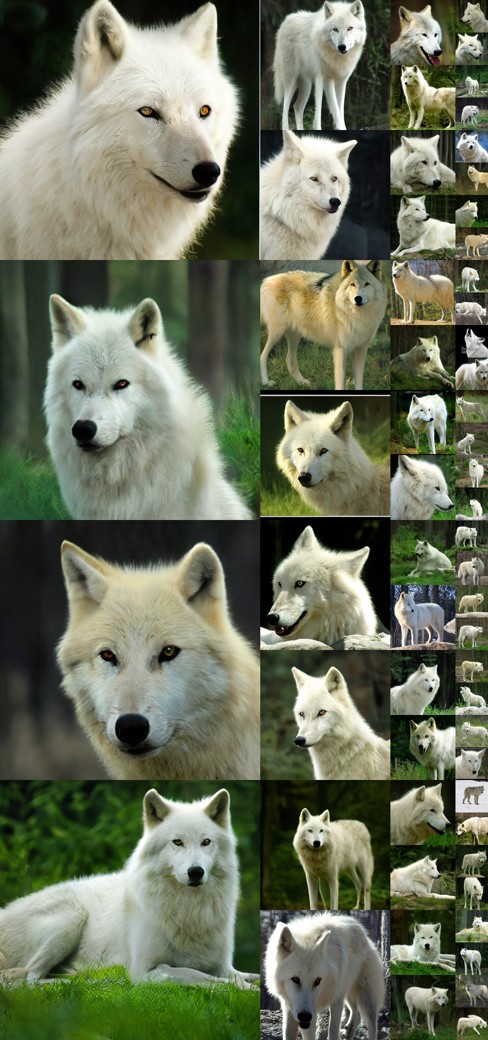}
    \caption*{Class label = ``arctic wolf'' (270)}
  \end{subfigure}
  \caption{\textbf{Uncurated 512$\times$512 FCDM-XL samples.} Classifier-free guidance scale = 4.0}
  \label{appendix_fig2}
\end{figure}

\begin{figure}[ht]
  \centering
  \begin{subfigure}[t]{0.49\columnwidth}
    \centering
    \includegraphics[width=\linewidth]{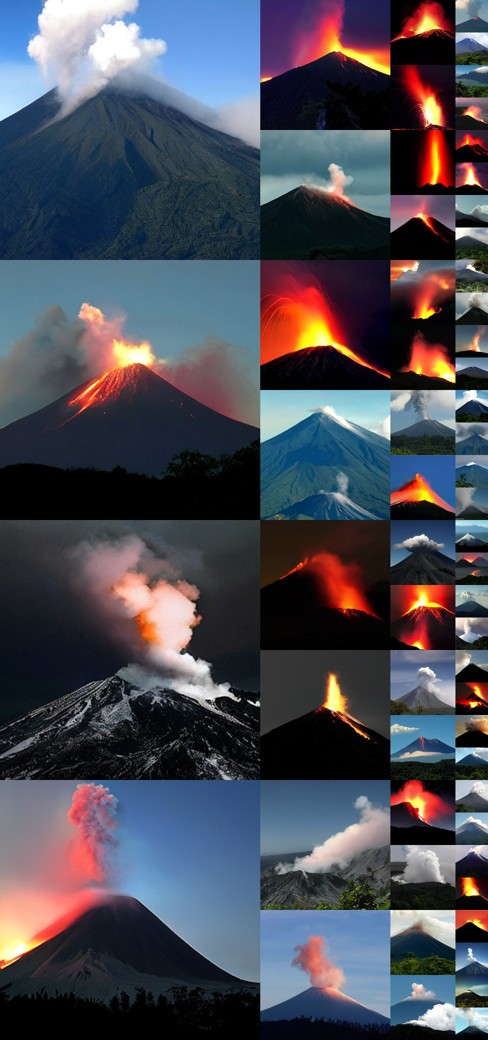}
    \caption*{Class label = ``volcano'' (980)}
  \end{subfigure}\hfill
  \begin{subfigure}[t]{0.49\columnwidth}
    \centering
    \includegraphics[width=\linewidth]{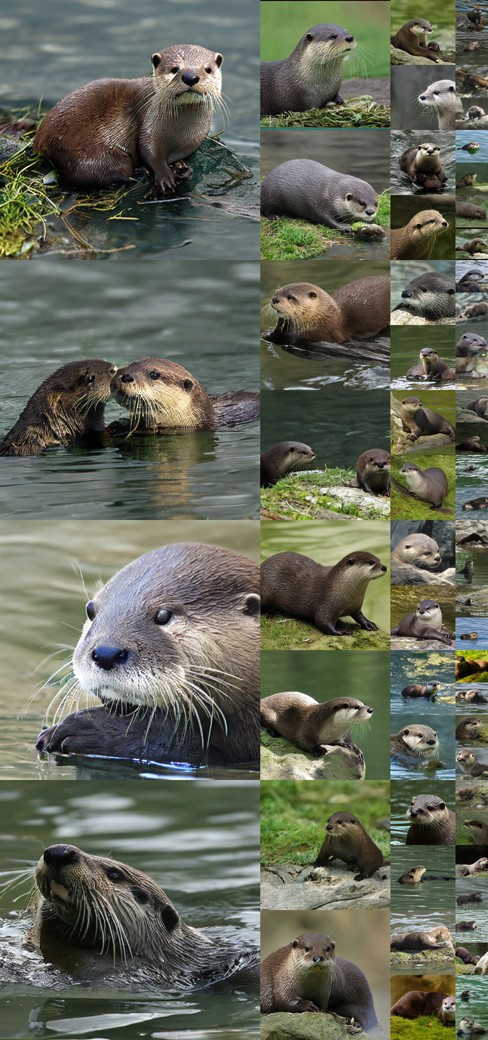}
    \caption*{Class label = ``otter'' (360)}
  \end{subfigure}
  \caption{\textbf{Uncurated 512$\times$512 FCDM-XL samples.} Classifier-free guidance scale = 4.0}
  \label{appendix_fig3}
\end{figure}

\begin{figure}[ht]
  \centering
  \begin{subfigure}[t]{0.49\columnwidth}
    \centering
    \includegraphics[width=\linewidth]{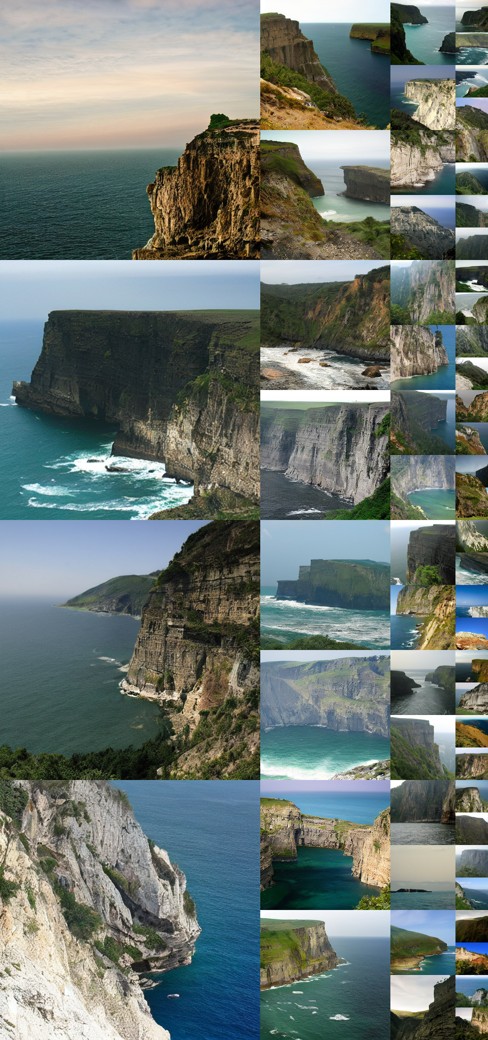}
    \caption*{Class label = ``cliff drop-off'' (972)}
  \end{subfigure}\hfill
  \begin{subfigure}[t]{0.49\columnwidth}
    \centering
    \includegraphics[width=\linewidth]{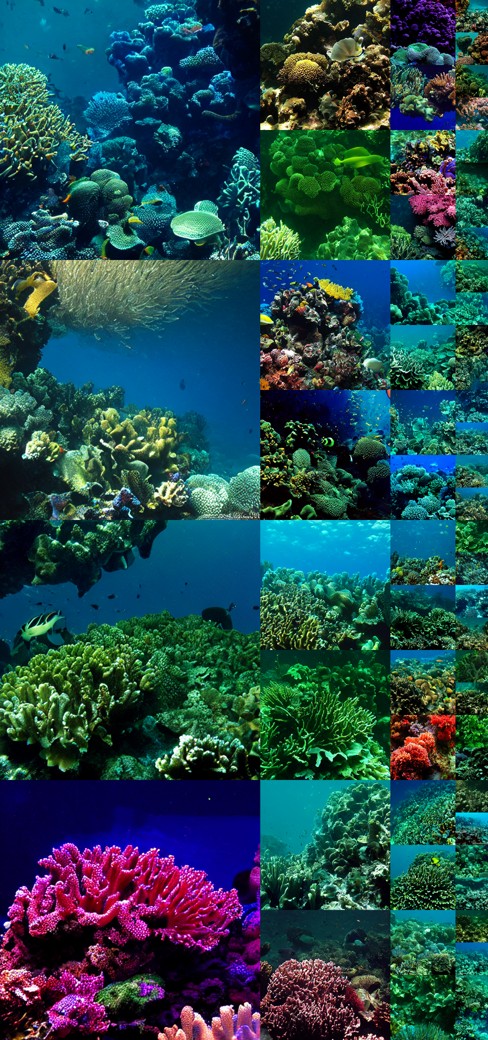}
    \caption*{Class label = ``coral reef'' (973)}
  \end{subfigure}
  \caption{\textbf{Uncurated 512$\times$512 FCDM-XL samples.} Classifier-free guidance scale = 4.0}
  \label{appendix_fig4}
\end{figure}

\begin{figure}[ht]
  \centering
  \begin{subfigure}[t]{0.49\columnwidth}
    \centering
    \includegraphics[width=\linewidth]{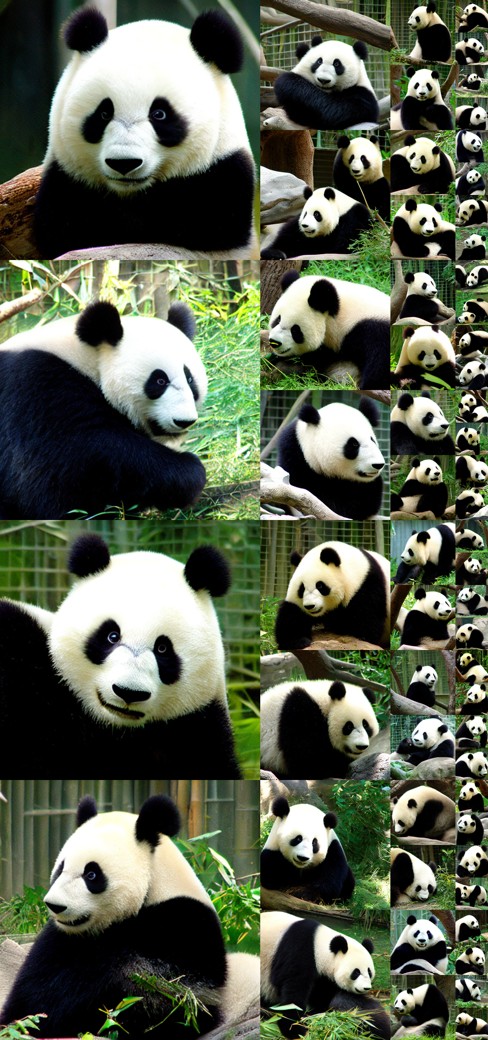}
    \caption*{Class label = ``panda'' (388)}
  \end{subfigure}\hfill
  \begin{subfigure}[t]{0.49\columnwidth}
    \centering
    \includegraphics[width=\linewidth]{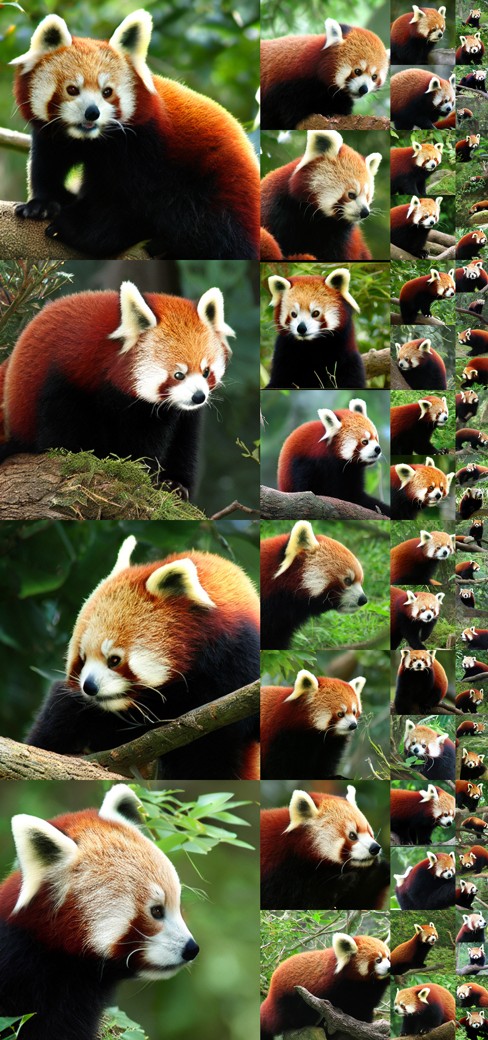}
    \caption*{Class label = ``red panda'' (387)}
  \end{subfigure}
  \caption{\textbf{Uncurated 512$\times$512 FCDM-XL samples.} Classifier-free guidance scale = 4.0}
  \label{appendix_fig5}
\end{figure}

\begin{figure}[ht]
  \centering
  \begin{subfigure}[t]{0.49\columnwidth}
    \centering
    \includegraphics[width=\linewidth]{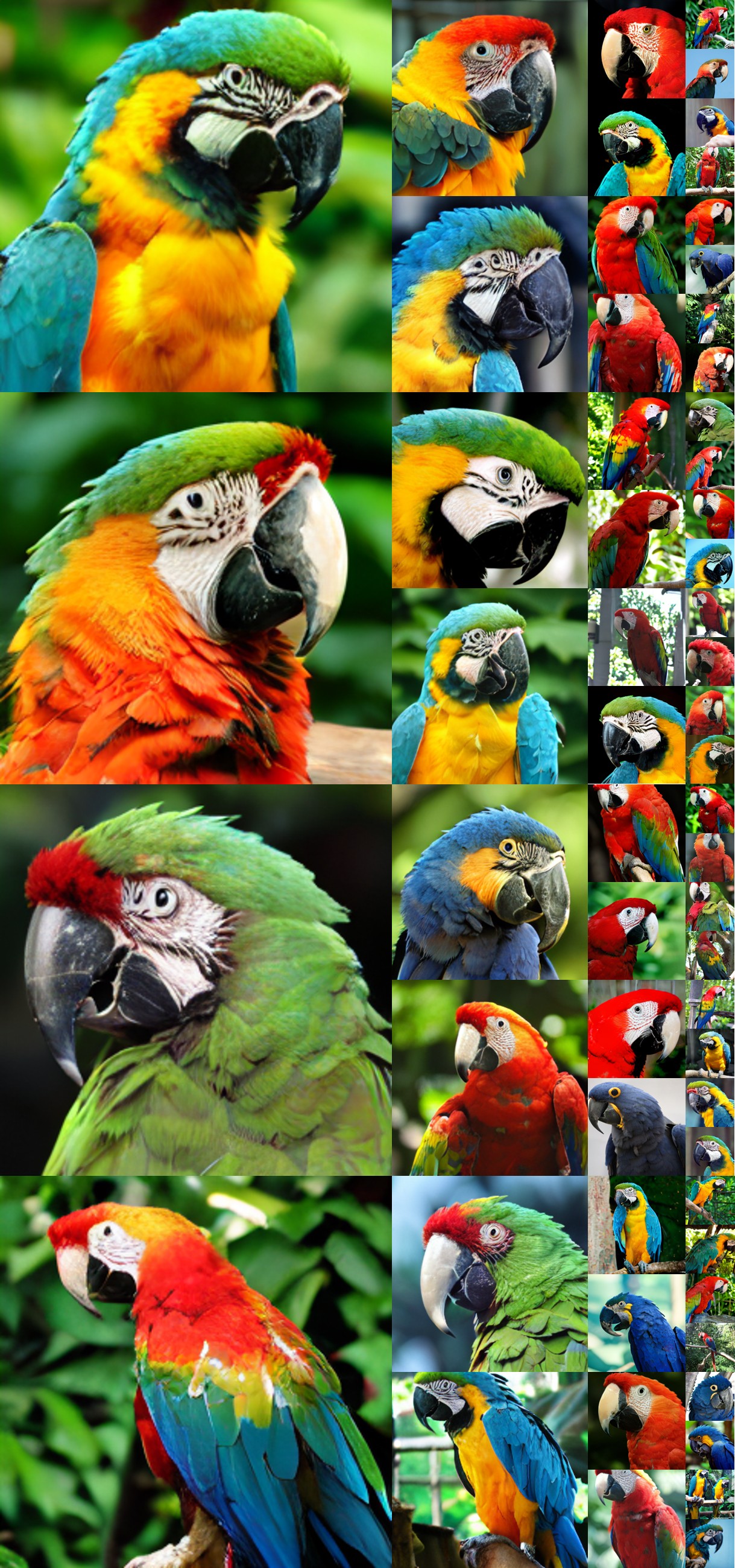}
    \caption*{Class label = ``macaw'' (88)}
  \end{subfigure}\hfill
  \begin{subfigure}[t]{0.49\columnwidth}
    \centering
    \includegraphics[width=\linewidth]{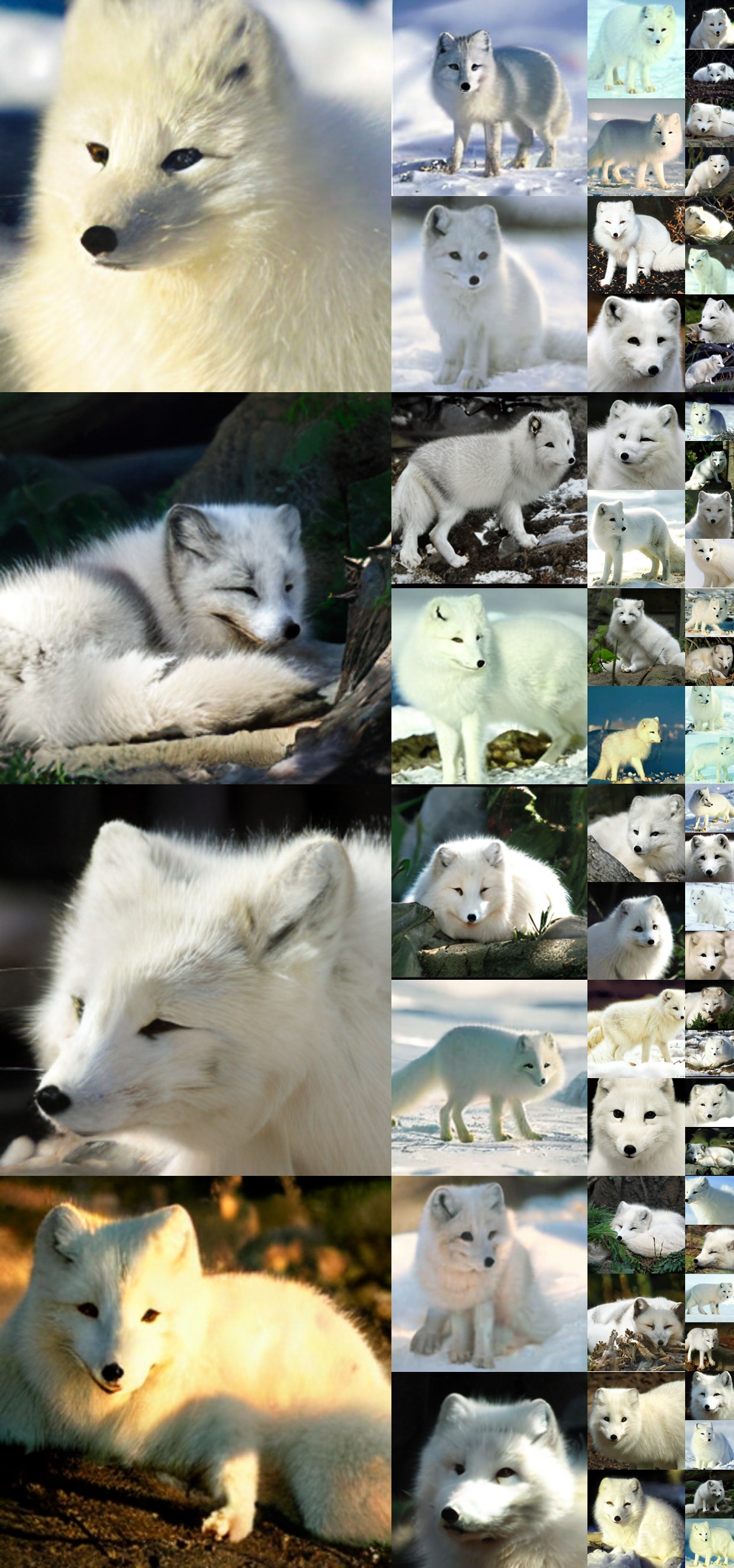}
    \caption*{Class label = ``arctic fox'' (279)}
  \end{subfigure}
  \caption{\textbf{Uncurated 256$\times$256 FCDM-XL samples.} Classifier-free guidance scale = 4.0}
  \label{appendix_fig6}
\end{figure}

\begin{figure}[ht]
  \centering
  \begin{subfigure}[t]{0.49\columnwidth}
    \centering
    \includegraphics[width=\linewidth]{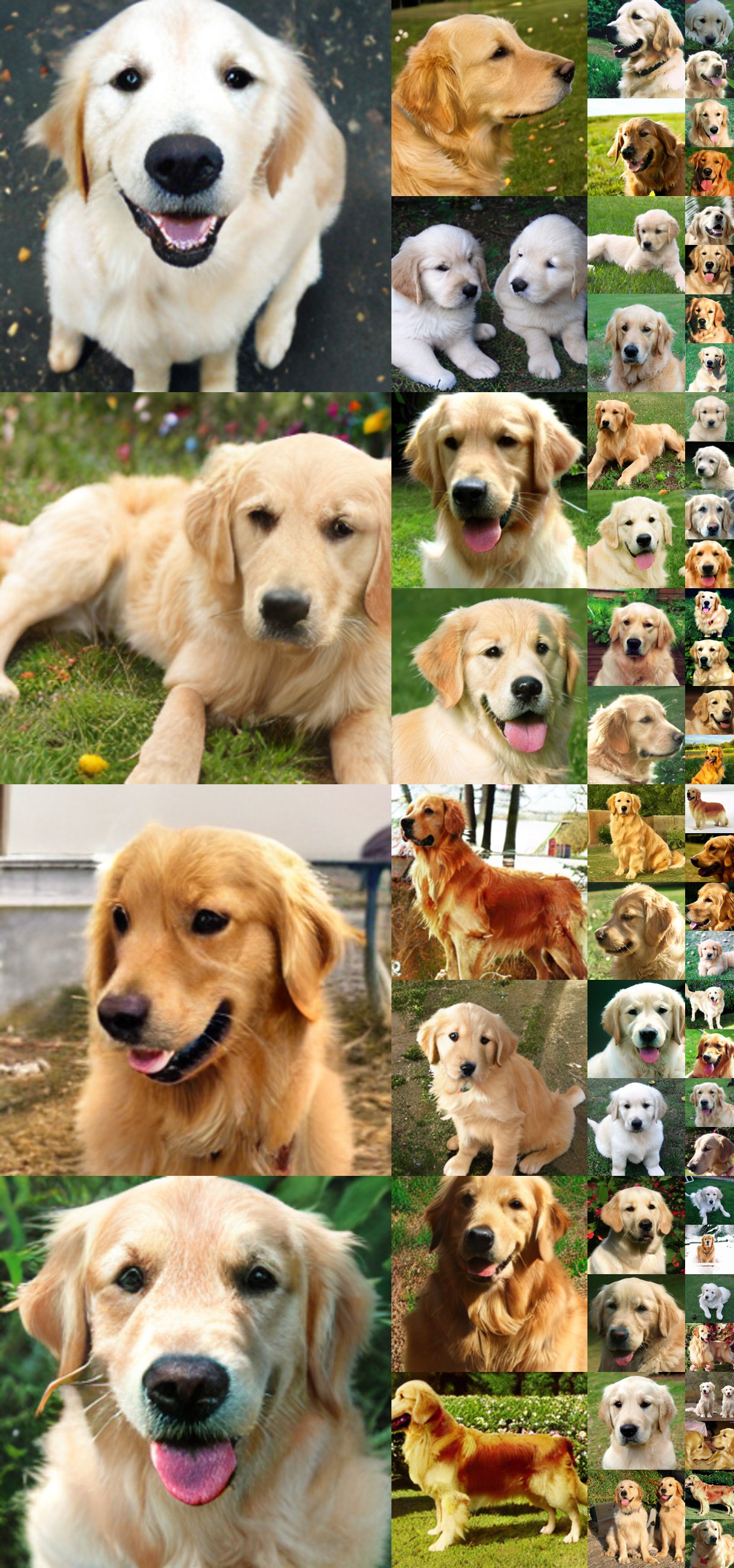}
    \caption*{Class label = ``golden retriever'' (207)}
  \end{subfigure}\hfill
  \begin{subfigure}[t]{0.49\columnwidth}
    \centering
    \includegraphics[width=\linewidth]{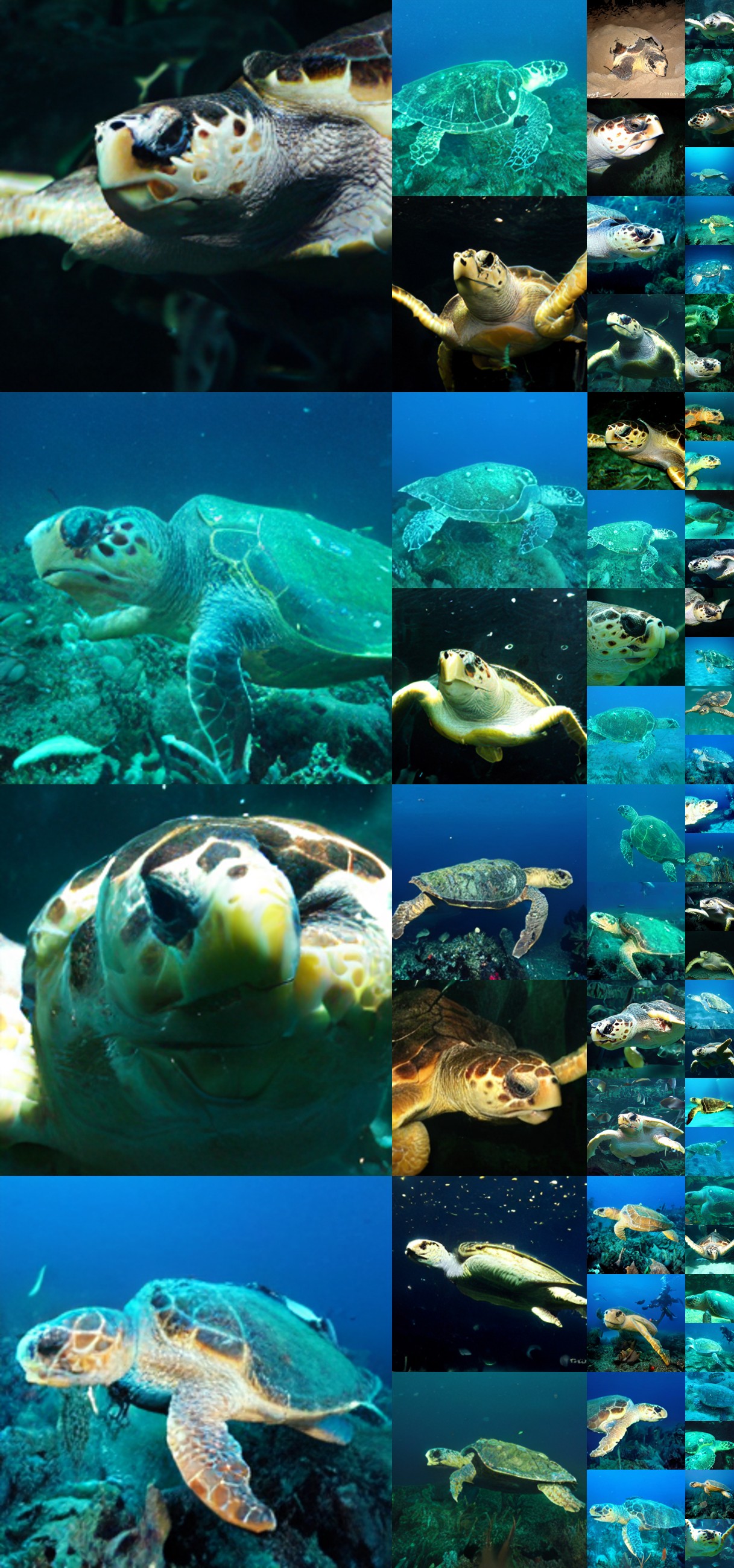}
    \caption*{Class label = ``loggerhead sea turtle'' (33)}
  \end{subfigure}
  \caption{\textbf{Uncurated 256$\times$256 FCDM-XL samples.} Classifier-free guidance scale = 4.0}
  \label{appendix_fig7}
\end{figure}

\begin{figure}[ht]
  \centering
  \begin{subfigure}[t]{0.49\columnwidth}
    \centering
    \includegraphics[width=\linewidth]{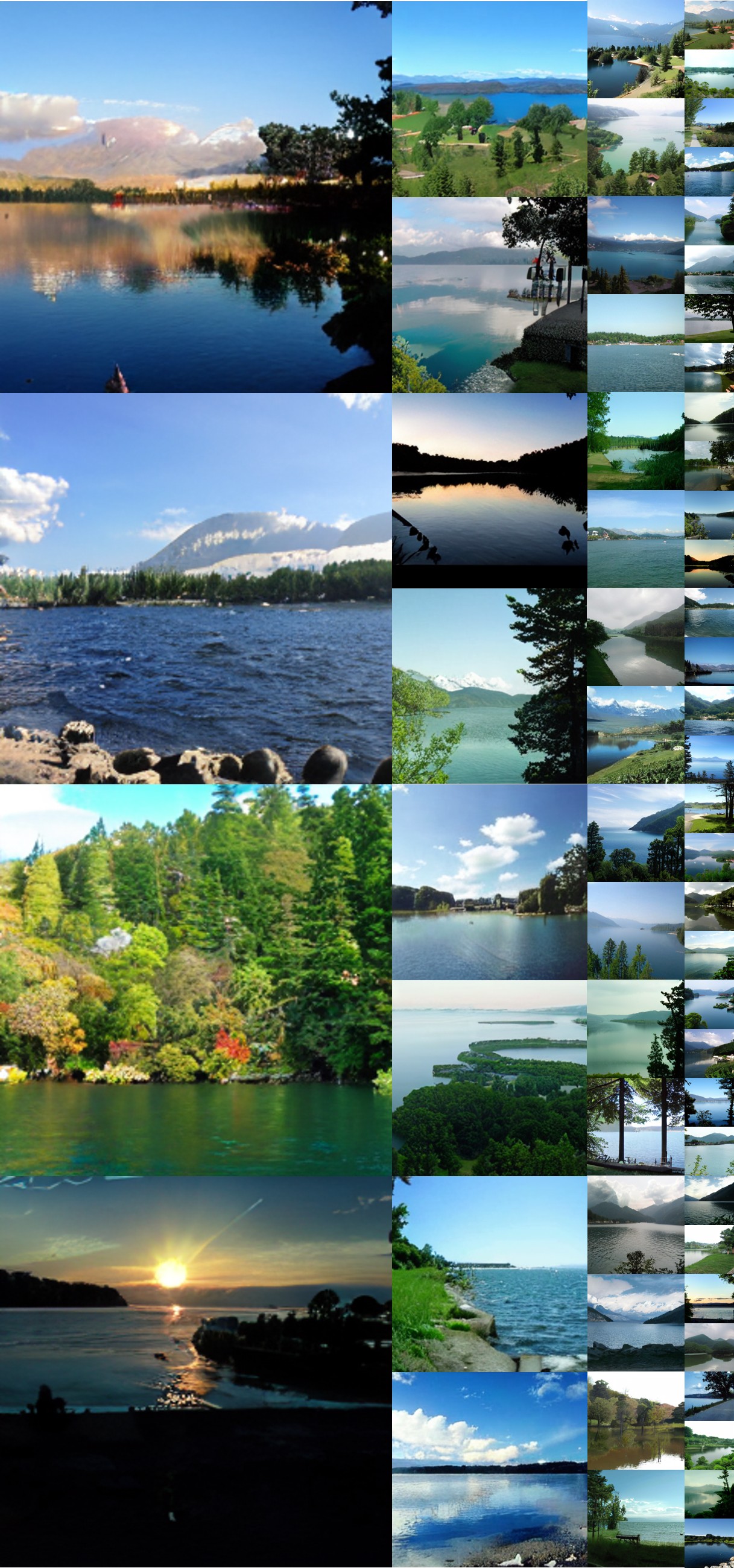}
    \caption*{Class label = ``lake shore'' (975)}
  \end{subfigure}\hfill
  \begin{subfigure}[t]{0.49\columnwidth}
    \centering
    \includegraphics[width=\linewidth]{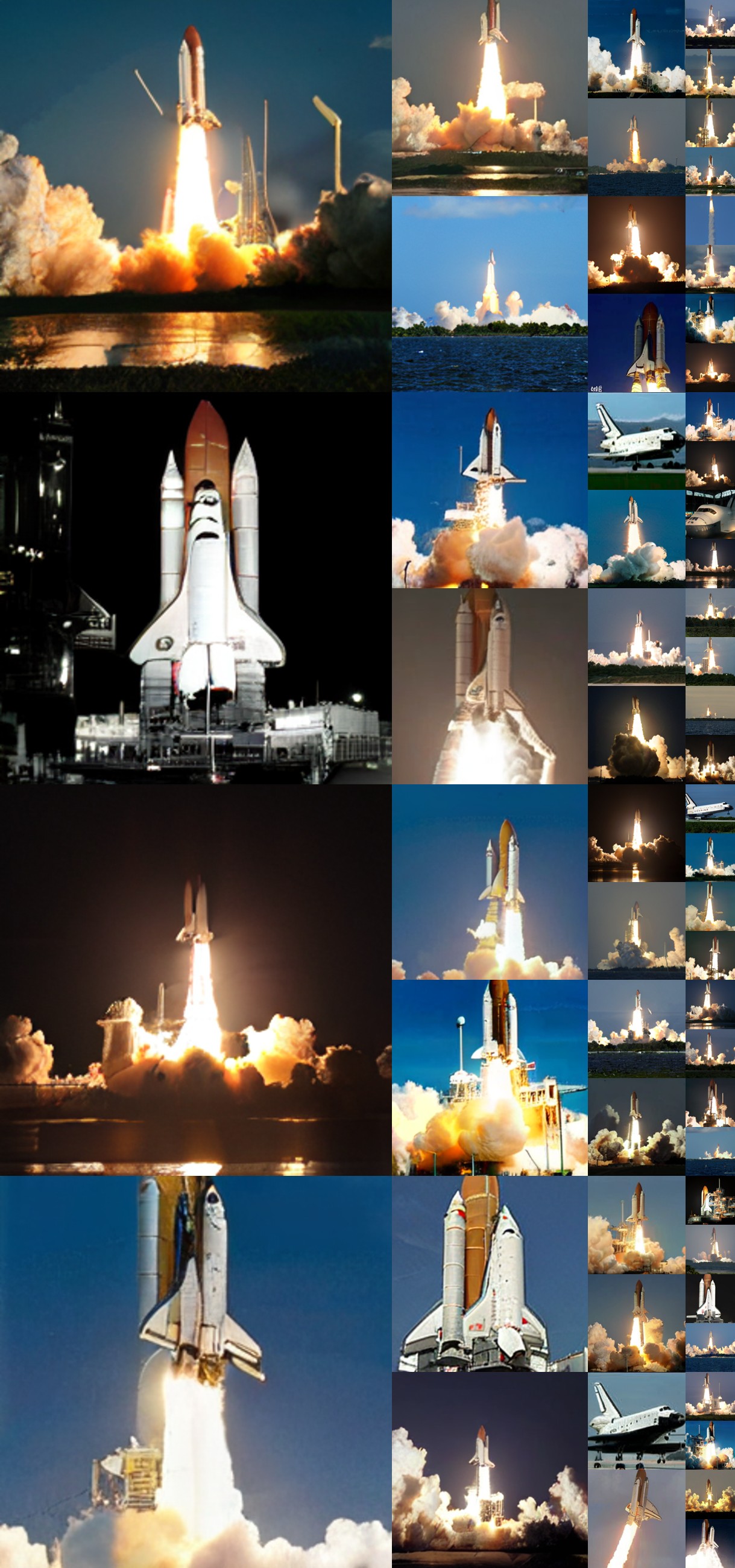}
    \caption*{Class label = ``space shuttle'' (812)}
  \end{subfigure}
  \caption{\textbf{Uncurated 256$\times$256 FCDM-XL samples.} Classifier-free guidance scale = 4.0}
  \label{appendix_fig8}
\end{figure}

\begin{figure}[ht]
  \centering
  \begin{subfigure}[t]{0.49\columnwidth}
    \centering
    \includegraphics[width=\linewidth]{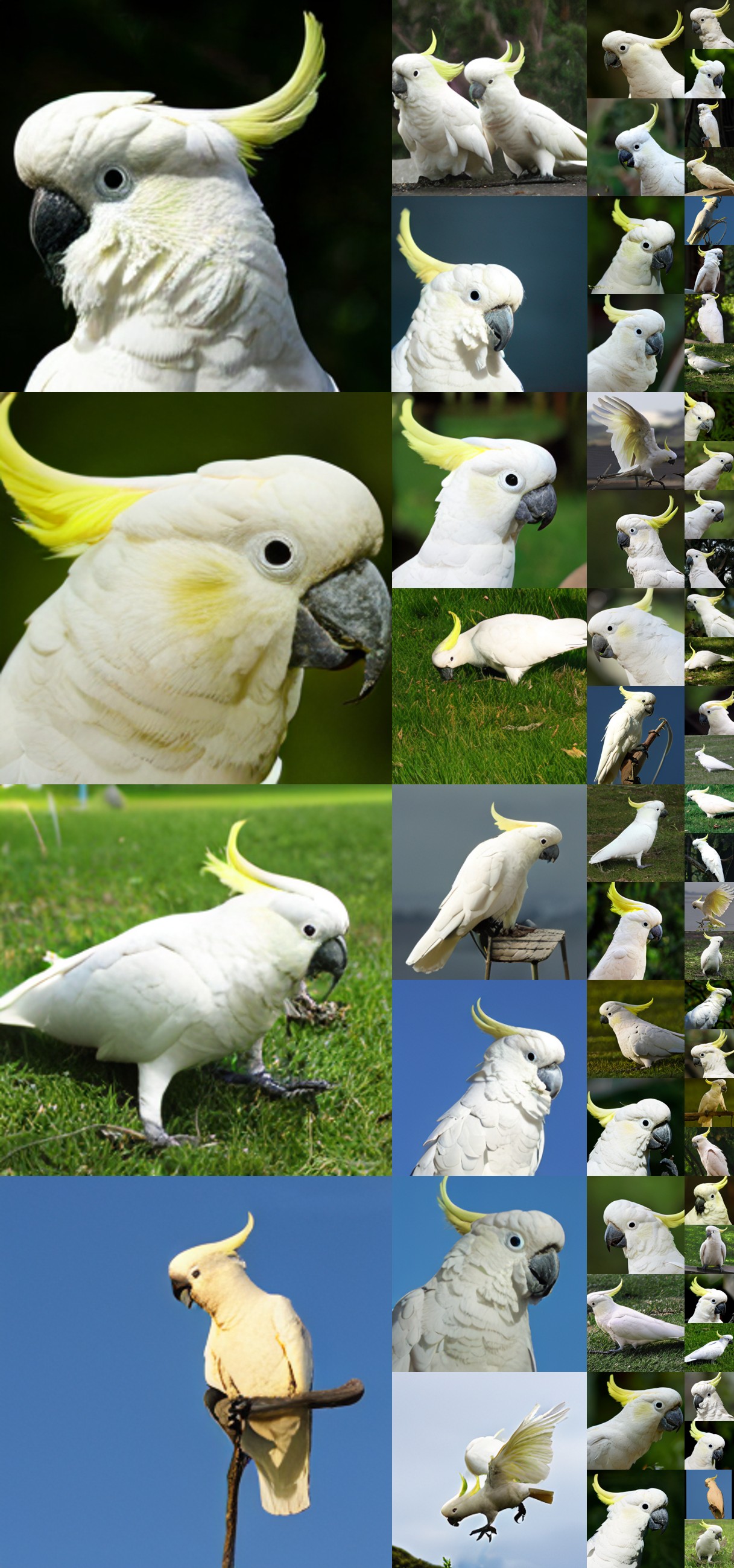}
    \caption*{Class label = ``sulphur-crested cockatoo'' (89)}
  \end{subfigure}\hfill
  \begin{subfigure}[t]{0.49\columnwidth}
    \centering
    \includegraphics[width=\linewidth]{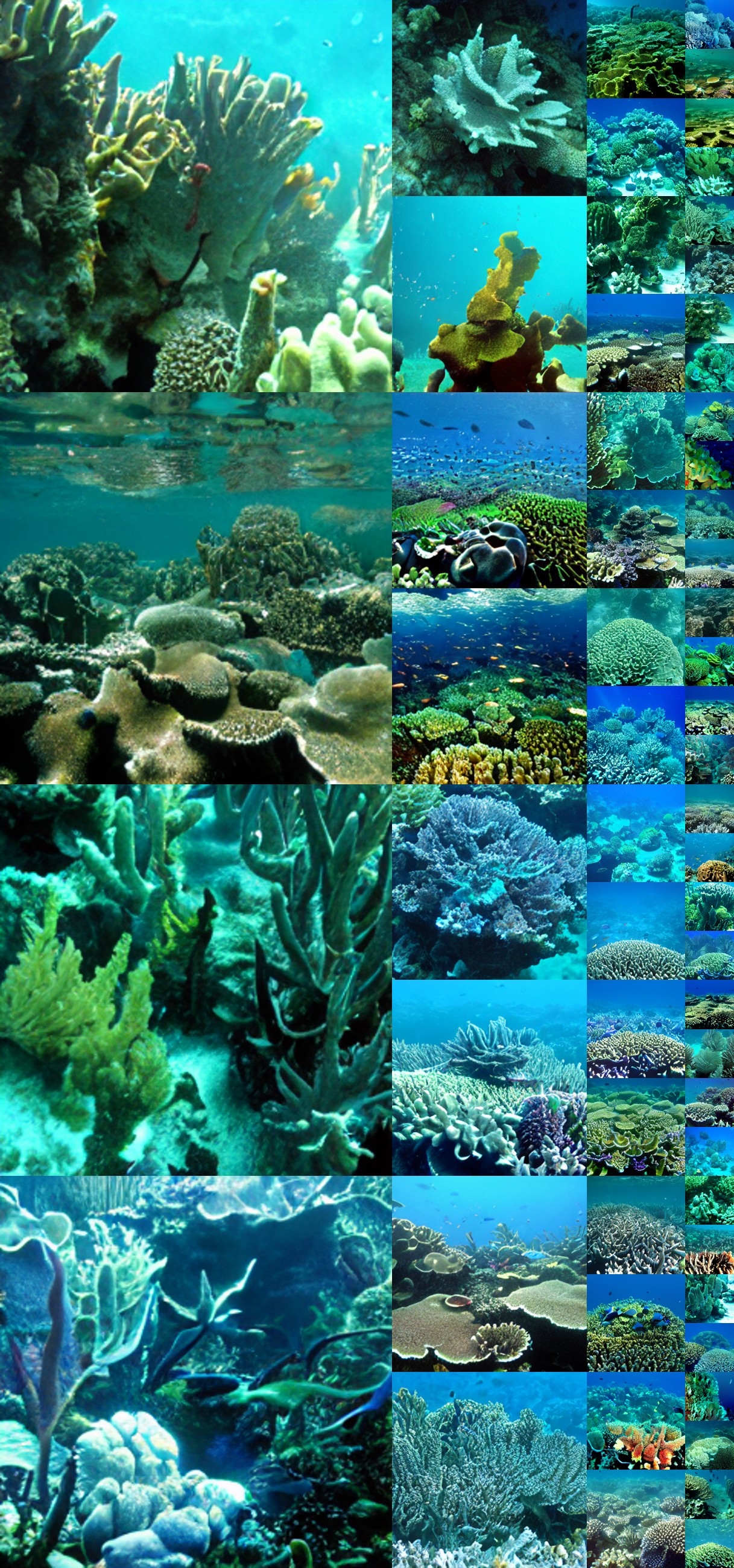}
    \caption*{Class label = ``coral reef'' (973)}
  \end{subfigure}
  \caption{\textbf{Uncurated 256$\times$256 FCDM-XL samples.} Classifier-free guidance scale = 4.0}
  \label{appendix_fig9}
\end{figure}

\begin{figure}[ht]
  \centering
  \begin{subfigure}[t]{0.49\columnwidth}
    \centering
    \includegraphics[width=\linewidth]{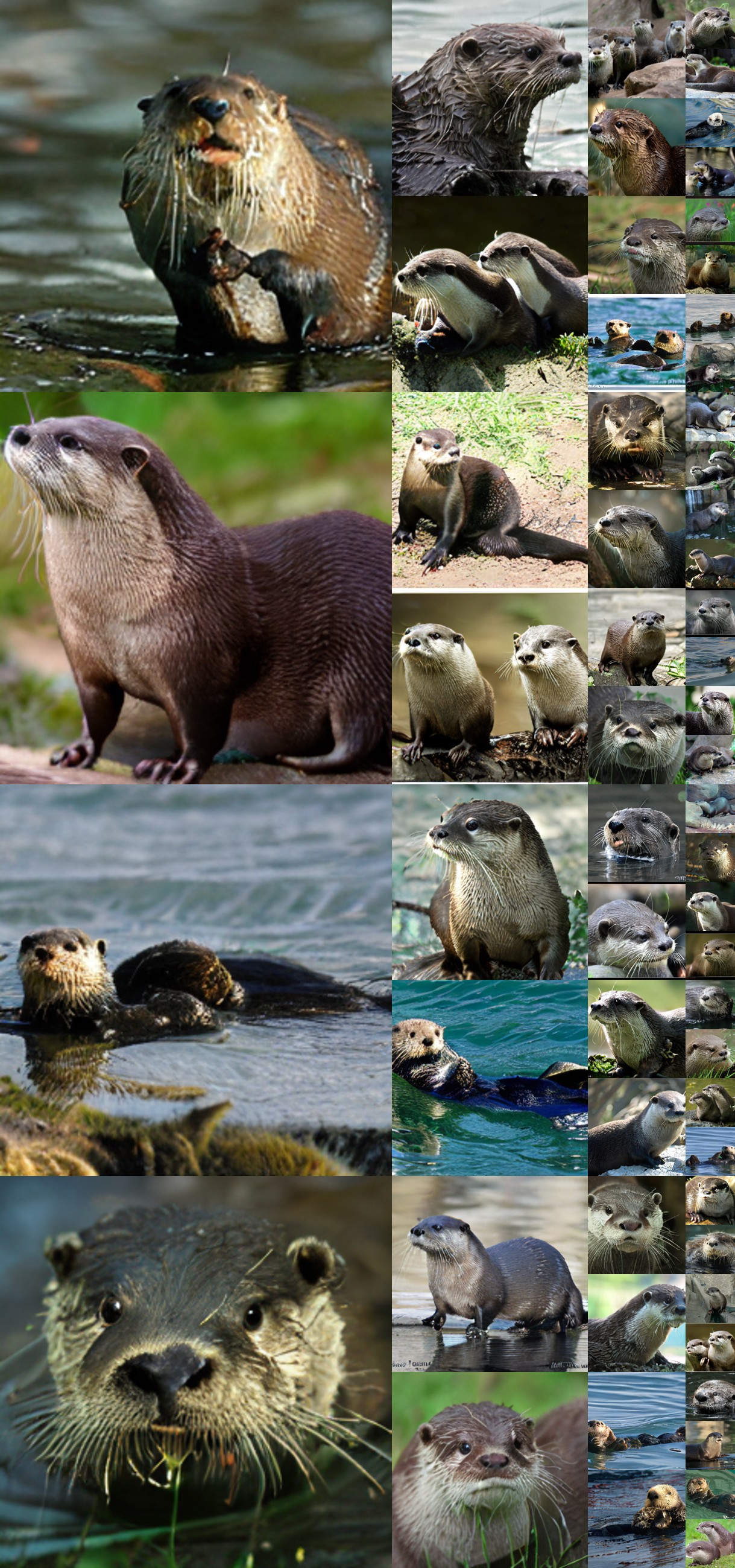}
    \caption*{Class label = ``otter'' (360)}
  \end{subfigure}\hfill
  \begin{subfigure}[t]{0.49\columnwidth}
    \centering
    \includegraphics[width=\linewidth]{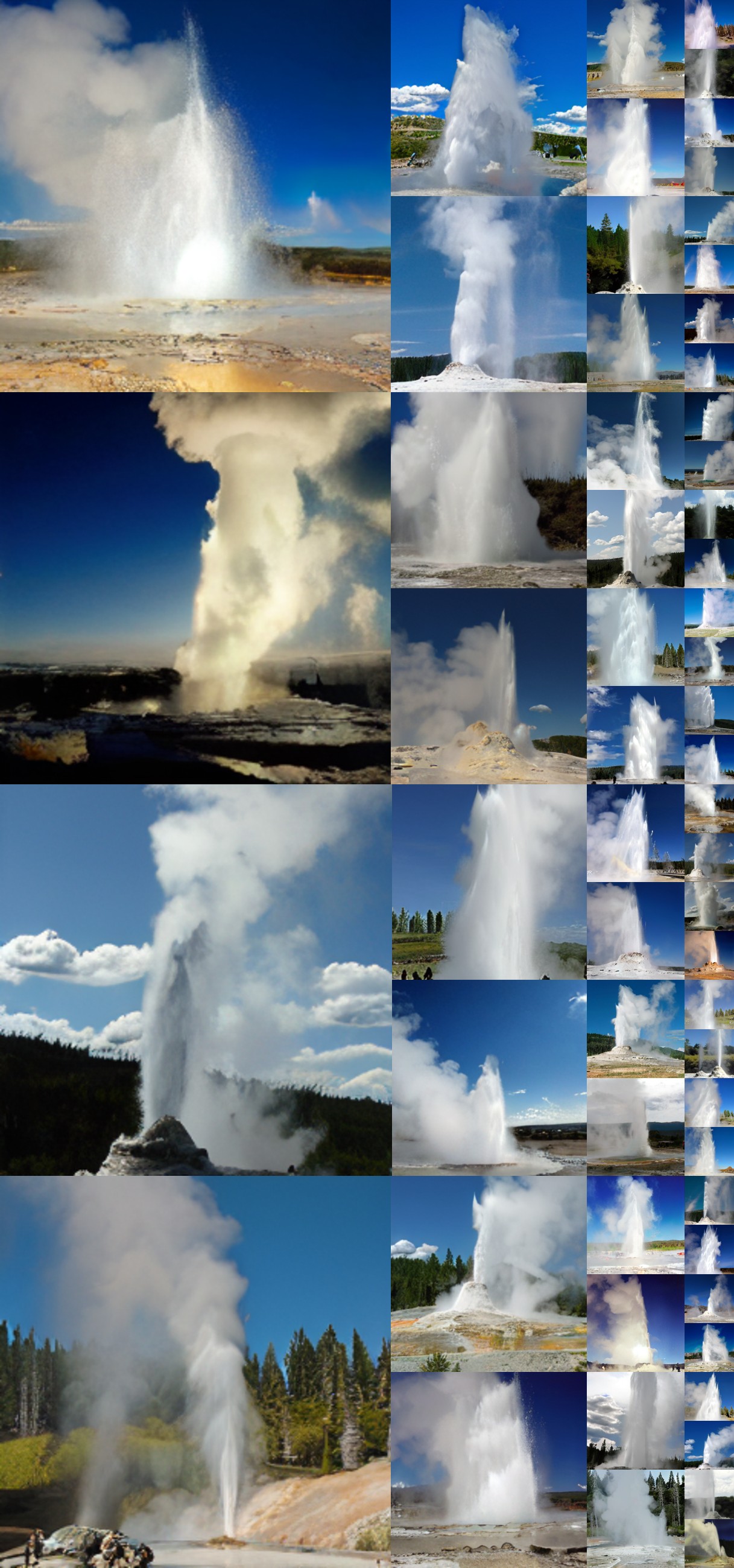}
    \caption*{Class label = ``geyser'' (974)}
  \end{subfigure}
  \caption{\textbf{Uncurated 256$\times$256 FCDM-XL samples.} Classifier-free guidance scale = 4.0}
  \label{appendix_fig10}
\end{figure}

\begin{figure}[ht]
  \centering
  \begin{subfigure}[t]{0.49\columnwidth}
    \centering
    \includegraphics[width=\linewidth]{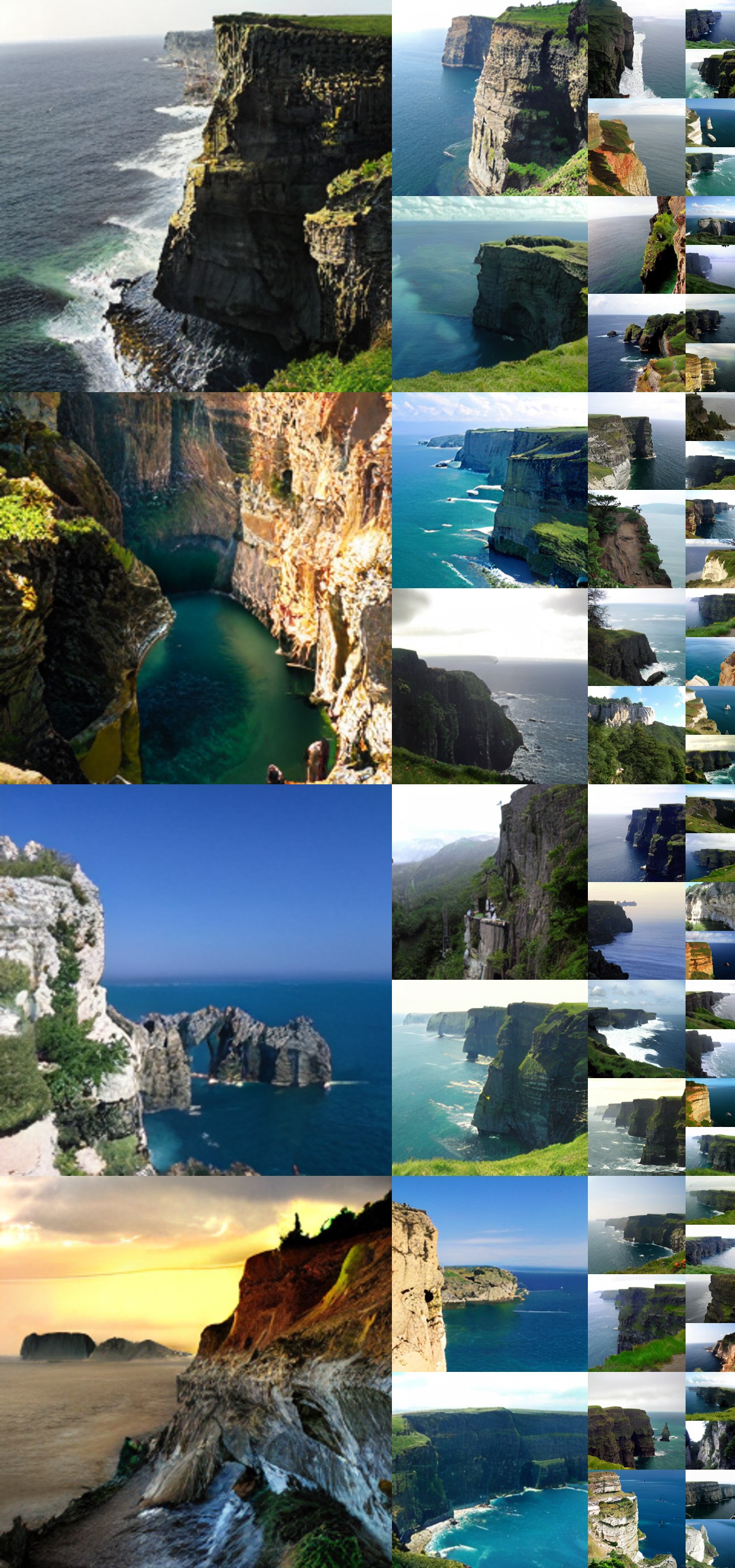}
    \caption*{Class label = ``cliff drop-off'' (972)}
  \end{subfigure}\hfill
  \begin{subfigure}[t]{0.49\columnwidth}
    \centering
    \includegraphics[width=\linewidth]{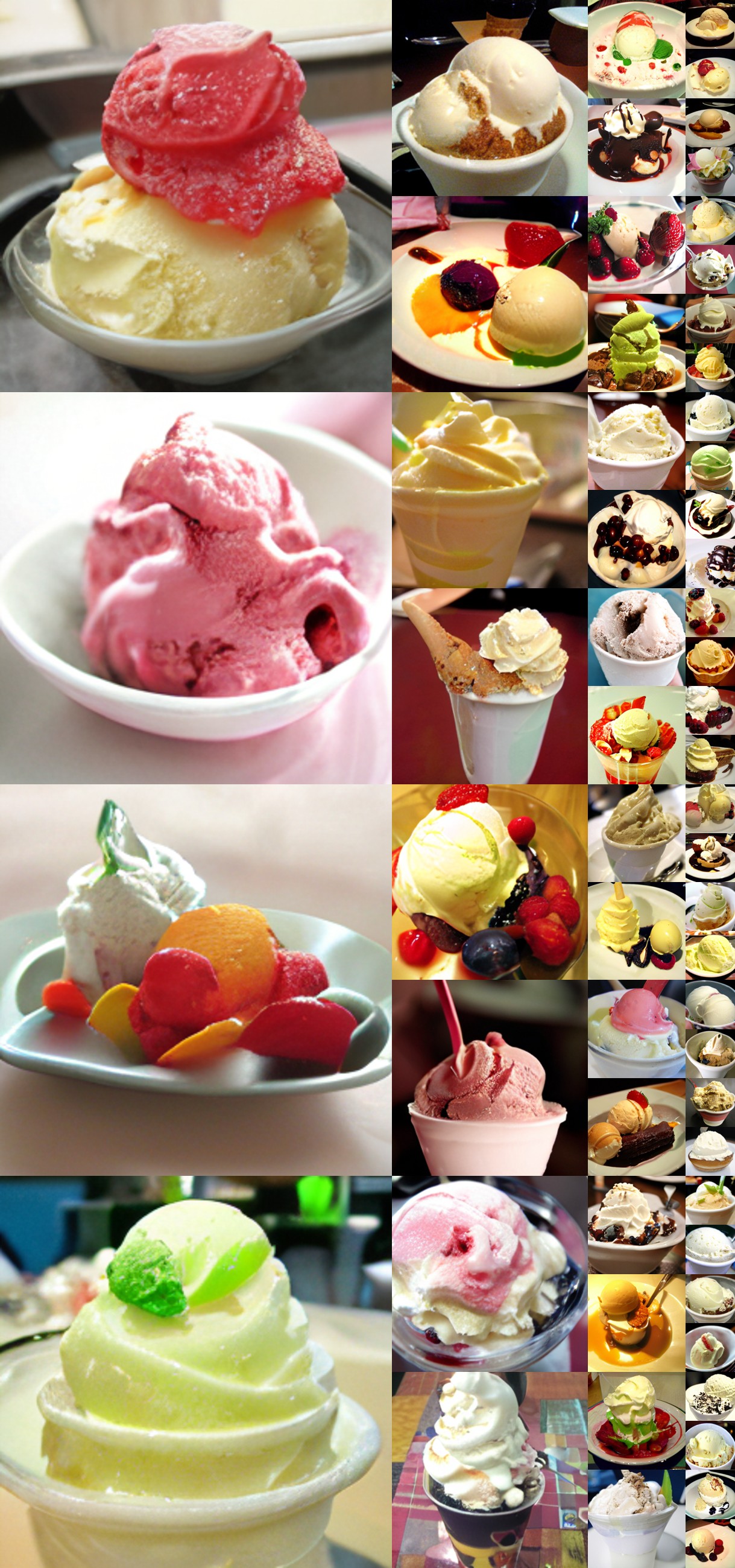}
    \caption*{Class label = ``ice cream'' (928)}
  \end{subfigure}
  \caption{\textbf{Uncurated 256$\times$256 FCDM-XL samples.} Classifier-free guidance scale = 4.0}
  \label{appendix_fig11}
\end{figure}

\end{document}